\documentclass[lettersize,journal]{IEEEtran}
\usepackage{amsmath,amsfonts}
\usepackage{algorithmic}
\usepackage{algorithm}

\usepackage[top=0.7in,bottom=0.5in,left=0.5in,textwidth=7.5in]{geometry}
\usepackage{array}
\usepackage[caption=false,font=normalsize,labelfont=sf,textfont=sf]{subfig}
\usepackage{textcomp}
\usepackage{stfloats}
\usepackage{url}
\usepackage{verbatim}
\usepackage{graphicx}
\usepackage{booktabs}       
\usepackage{multirow}
\usepackage{cite}
\usepackage{color}
\usepackage{bbding}
\usepackage{ulem}
\usepackage{colortbl}
\usepackage[table]{xcolor}
\newcommand{\etal}{\textit{et al}.}
\newcommand{\ie}{\textit{i}.\textit{e}.}
\newcommand{\eg}{\textit{e}.\textit{g}.}

\usepackage[pagebackref=false,breaklinks=true,linkcolor=red,anchorcolor=blue,            citecolor=green,letterpaper=true,colorlinks,bookmarks=false]{hyperref}

\begin{document}

\title{Enhancing Information Maximization with Distance-Aware Contrastive Learning for \\ Source-Free Cross-Domain Few-Shot Learning}

\author{Huali Xu, Li Liu, Shuaifeng Zhi, Shaojing Fu,  Zhuo Su, Ming-Ming Cheng, Yongxiang Liu
\thanks{This work was partially supported by National Key Research and Development Program of China No. 2021YFB3100800, the Academy of Finland under grant 331883, Infotech Project FRAGES,
the National Natural Science Foundation of China under Grant 62376283 and the Key Stone grant (JS2023-03) of the National University of Defense Technology (NUDT). Corresponding authors: Li Liu and Yongxiang Liu.}
\thanks{Huali Xu and Zhuo Su are with the Center for Machine Vision and Signal Analysis (CMVS), University of Oulu, Oulu 90570, Finland (email: huali.xu@oulu.fi; zhuo.su@oulu.fi).
Shuaifeng Zhi, Yongxiang Liu, and Li Liu are with the College of Electronic Science and Technology, National University of Defense Technology, Changsha, 410073, China (email: zhishuaifeng@outlook.com; lyx\_bible@sina.com; lilyliu\_nudt@163.com). Shaojing Fu (fushaojing@nudt.edu.cn) is with the College of Computer, National University of Defense Technology.
Ming-Ming Cheng (cmm@nankai.edu.cn) is with the College of Computer Science, Nankai University, Tianjin, TKLNDST, China.}}

\markboth{Submission to IEEE Transactions on Image Processing}%
{How to Use the IEEEtran \LaTeX \ Templates}


\maketitle

\begin{abstract}
Existing Cross-Domain Few-Shot Learning (CDFSL) methods require access to source domain data to train a model in the pre-training phase. However, due to increasing concerns about data privacy and the desire to reduce data transmission and training costs, it is necessary to develop a CDFSL solution without accessing source data. For this reason, this paper explores a Source-Free CDFSL (SF-CDFSL) problem, in which CDFSL is addressed through the use of existing pretrained models instead of training a model with source data, avoiding accessing source data. However, due to the lack of source data, we face two key challenges: effectively tackling CDFSL with limited labeled target samples, and the impossibility of addressing domain disparities by aligning source and target domain distributions. This paper proposes an Enhanced Information Maximization with Distance-Aware Contrastive Learning (IM-DCL) method to address these challenges. Firstly, we introduce the transductive mechanism for learning the query set. Secondly, information maximization (IM) is explored to map target samples into both individual certainty and global diversity predictions, helping the source model better fit the target data distribution. However, IM fails to learn the decision boundary of the target task. This motivates us to introduce a novel approach called Distance-Aware Contrastive Learning (DCL), \textcolor{black}{in which we consider the entire feature set as both positive and negative sets, akin to Schrödinger's concept of a dual state. Instead of a rigid separation between positive and negative sets, we employ a weighted distance calculation among features to establish a soft classification of the positive and negative sets for the entire feature set. We explore three types of negative weights to enhance the performance of CDFSL. Furthermore, we address issues related to IM by incorporating contrastive constraints between object features and their corresponding positive and negative sets.} Evaluations of the 4 datasets in the BSCD-FSL benchmark indicate that the proposed IM-DCL, without accessing the source domain, demonstrates superiority over existing methods, especially in the distant domain task. Additionally, the ablation study and performance analysis confirmed the ability of IM-DCL to handle SF-CDFSL. The code will be made public at \url{https://github.com/xuhuali-mxj/IM-DCL}.
\end{abstract}


\section{Introduction}
Humans can rapidly learn novel concepts with several examples using prior knowledge~\cite{lake2015human}, \ie, Few-Shot Learning (FSL). Inspired by how humans learn, recently, FSL has emerged as a hot subfield of machine learning. Machines address a target task with only several examples (\eg, an object classification task with $N$ classes, each of which has only several example images), usually with the assistance of a target task-irrelevant source dataset (\eg, a dataset of object images not from the target classes) that is typically large scale and labeled~\cite{wang2020generalizing,sun2023fastal,song2023comprehensive,yang2022efficient,vu2023instance,guo2022learning}. The benefits of studying FSL are threefold. Firstly, the machine can solve the task with only a few samples, which makes its intelligence closer to that of humans. Secondly, FSL can train models for infrequent scenarios when it is challenging or impossible to acquire enough supervised data. FSL algorithms reduce data gathering and computation costs. Due to its significance and various applications, FSL has been widely studied recently and numerous approaches have been proposed, with excellent surveys~\cite{wang2020generalizing,song2023comprehensive}.


Despite recent progress, conventional FSL assumes that the auxiliary source dataset and the target dataset are from the same domain~\cite{wang2020generalizing,song2023comprehensive}. This has serious limitations for some realistic applications such as medical image analysis and remote sensing image analysis where obtaining a large source dataset from the same domain is difficult or even impossible. To mitigate this limitation, Cross Domain FSL, using auxiliary data from other domains to assist the learning of the target task, has become a subfield of FSL~\cite{xu2023deep}. Certainly, due to the domain gap between the source and target domains, CDFSL poses even greater challenges than FSL.

In the standard CDFSL setting, the FSL task in the target domain is solved by taking advantage of data from other domains, which means that the source dataset could be accessed while learning the target task. Existing CDFSL methods~\cite{tseng2020cross,phoo2020self,islam2021dynamic,xu2022cross,fu2023styleadv,zhao2023fs} emphasize the acquisition of shared knowledge within the source domain to improve FSL accuracy in the target domain. 
Other technologies solve CDFSL by exploring adaptation algorithms in the target domain~\cite{li2022ranking,zhao2023dual}.
However, these methods are not realistic in many real-world scenarios~\cite{liang2020we}. Firstly, it is computationally expensive to train with a large source dataset, especially for edge devices. Secondly, due to confidentiality, privacy, and copyright concerns, the source dataset might not be available. Lastly, the current foundational models (FMs), known for their robust generalization capabilities, has facilitated their broad application in diverse downstream tasks, all without the need for access to source domain and fails to employ the training strategies designed for CDFSL.

\begin{figure*}[!t]
  \centering
  \includegraphics[width=\textwidth]{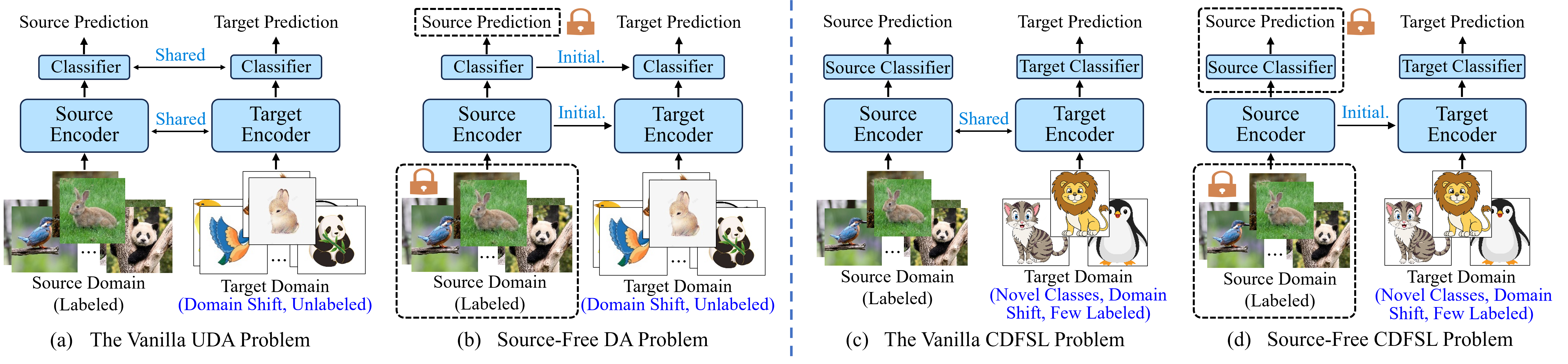}
  \caption{Contrast the proposed SF-CDFSL problem with existing closely related problem settings. (a) Vanilla UDA: Both source and target tasks share the same label space, but the data of these two domains are from different distributions. (b) SFDA: the source data are not accessible in UDA. Contrasted with the DA problem, (c) Vanilla CDFSL: including sufficient labeled source data and a few labeled target data (there are both domain gaps and task shift between source and target task). While (d) Our proposed SF-CDFSL: in which the source data is not accessible (like SFDA) and contains both domain gaps and task shift (like CDFSL), is more challenging. In SF-CDFSL, we solve the FSL problem in the target domain with limited target data and a pretrained source model without accessing any source data. This effectively eliminates the potential information leakage from the source domain. The lock means that the corresponding part is not accessible when retraining the target task, accessible only for pre-training.}
  \label{sfcdfsl}
  \vspace{-0.3cm}
\end{figure*}

Therefore, to mitigate the dependence of CDFSL methods on the source dataset, motivated by the well-established problem of Source-Free Domain Adaption (SFDA)~\cite{liang2020we,ding2022source}, we propose to address a novel yet highly challenging problem of Source-Free CDFSL (SF-CDFSL). To our best knowledge, we are the first to provide a timely investigation of this scenario. In SF-CDFSL, our objective is to learn the FSL task in the target domain only using an existing pretrained model (called source model) and without having access to any source data (as shown in Figure~\ref{sfcdfsl} (d), contrasting with three other closely related problems). Since many pretrained foundation models~\cite{radford2021learning,dosovitskiy2020image,liu2020feature} are available now, we believe the problem of SF-CDFSL will attract increasing attention. \textcolor{black}{Despite its great significance, finding solutions to this realistic SF-CDFSL problem is quite challenging due to the following two reasons: (1) Different from CDFSL, SF-CDFSL uses only limited labeled target data to solve the FSL problem in the target domain. (2) It is impossible to resolve domain gaps by aligning the source and target distributions since the source data distribution is unknown.}

To address the inherent challenges in SF-CDFSL and delve deeper into the topic, this paper introduces a novel framework named Enhanced Information Maximization with Distance-Aware Contrastive Learning (IM-DCL). For the first challenge, we acknowledge that there are only a few labeled samples available in the target domain, so we consider all samples (both labeled support and unlabeled query sets). For this, instead of the traditional ``specific-to-general-to-specific'' inductive learning approach typically employed in CDFSL solutions, we introduce transductive learning~\cite{vapnik1999nature} during the learning phase, which incorporates the query set of the target domain into the adaptation process, effectively achieving the transition from ``specific-to-specific''~\cite{gammerman2013learning}. To address the second challenge, we introduce information maximization to directly constrain the model to produce prediction results of individual certainty and general diversity. On the one hand, this means that the source model can handle the target task without aligning the source and target distributions. On the other hand, IM does not need supervised information, which provides the support for exploiting query sets under the transduction mechanism. 

However, the goal of IM is to use the source model to produce satisfactory predictions for the target samples, rather than strictly insisting on correct class assignments. For instance, even if features are incorrectly, uniformly, and explicitly classified in wrong categories, IM still provides correct feedback. Therefore, to ensure that the IM contains the correct category distribution, \textcolor{black}{Existing CL methods use strict labels to classify positive and negative sets,. However, when dealing with unlabeled query sets, this label-based approach becomes impractical. To tackle this issue, we introduce Distance-Aware Contrastive Learning (DCL), a method that softly classifies positive and negative sets based on feature distances. In DCL, we pioneer the use of Schrödinger's concept in contrastive learning, where both positive and negative sets cover the entire feature set. Unlike traditional methods that rely on strict labels, DCL employs feature distances for more adaptable weighting and classification. The process starts by extracting all features from the memory bank, containing all target domain features. Next, we compute distances between each feature and the object feature to determine their weights. Features close to the object receive higher weights, while distant ones receive lower weights. These weighted values are then combined with corresponding features in the memory database to create positive sets for object features. Conversely, the negative set is formed by inversely assigning weights, \ie, features farther apart receive greater weights, while closer ones receive smaller weights. We experiment with three weight inversion methods: reverse order, opposite, and non-linear mapping. DCL's advantage lies in improving FSL performance on the target domain by reducing the distance between object features and their positive set counterparts while increasing their distance from negative set features. This approach effectively addresses classification alignment challenges associated with IM loss. The details are provided in Section~\ref{mnpc}.}
In general, the key contributions to this paper can be summarized as follows:
\begin{itemize}
\item{This paper introduces the novel Source-Free Cross-Domain Few-Shot Learning (SF-CDFSL) problem, aiming to address target tasks without source data access and training. SF-CDFSL minimizes reliance on the source domain and pre-training, solving CDFSL tasks efficiently, ensuring source privacy, and reducing computation and transfer costs.}
\item{We introduce Enhanced Information Maximization with Distance-Aware Contrastive Learning (IM-DCL) to tackle the challenges of SF-CDFSL. IM-DCL combines transductive learning, infuses information maximization (IM) for confident and diverse results, and enhances decision boundary learning via Distance-Aware Contrastive Learning (DCL).}
\item{Comprehensive experiments on various source models and the BSCD-FSL benchmark demonstrate the state-of-the-art performance of the proposed IM-DCL. IM-DCL excels against adapted strategy-driven methods and holds competitive ground against training strategy-based approches, particularly in distant domains.}
\end{itemize}

\section{Related work}
\subsection{Cross-Domain Few-Shot Learning}
\textcolor{black}{Cross-Domain Few-Shot Learning (CDFSL) is a novel techniques for solving FSL challenges in the target domain, which supportted by the source task with the sufficient source data (source and target data are from different domain).} Since Guo \etal~\cite{guo2020broader} set the BSCD-FSL benchmark and Tseng \etal~\cite{tseng2020cross} conceptualized it, many studies have appeared in CDFSL~\cite{sun2021explanation,phoo2020self}. Most of them focused on pre-training strategies in CDFSL~\cite{phoo2020self,islam2021dynamic,xu2022cross,fu2021meta}. A common practice is to introduce additional data during the training phase to facilitate the acquisition of shared knowledge between the source and the target domain~\cite{phoo2020self,islam2021dynamic,fu2022generalized,fu2021meta}. \textcolor{black}{For example,~\cite{phoo2020self} learns the source domain representation with a self-training strategy by using unlabeled target data. Following the above method,~\cite{islam2021dynamic} proposes a dynamic distillation-based approach, utilizing unlabeled images from the target domain. Also,~\cite{fu2021meta} uses a small amount of labeled target data for training. They introduced the meta-FDMixup network, which helps the model by mixing up different data and separating important features.} In addition, other CDFSL solutions incorporate additional information, such as style, during the pre-training phase~\cite{xu2022cross,fu2022wave,zhang2022free,fu2023styleadv}. \textcolor{black}{~\cite{xu2022cross} introduces an ISSNet, which enhances model generalization by applying styles from unlabeled data to labeled data. ~\cite{fu2022wave} tackle CDFSL by analyzing style variations, breaking down images into simpler components using wavelet transform to better understand their styles. Furthermore, they also develop a training method called StyleAdv, focusing on meta style adversarial training for CDFSL. Besides,~\cite{zhang2022free} proposes the SET-RCL method, which adapts to new styles by considering both data and model adjustments to mimic the styles of unknown domains. Additionally, ~\cite{graph1} introduces a Gia-CFSL framework, combining few-shot learning with domain alignment in hyperspectral image classification through graph information, to bridge the gap between different domains effectively.} Currectly, CDFSL technology focus on adaptation strategies in the target domain. This is in addition to in-depth analyses of training strategies in the source domain~\cite{li2022ranking,zhao2023dual}. \textcolor{black}{~\cite{li2022ranking} use traditional distance-based classifiers and image retrieval views, they apply a reranking procedure to refine the target distance matrix by identifying k-reciprocal neighbors in the task.~\cite{zhao2023dual} proposes a dual adaptive representation alignment approach that leverages prototypical feature alignment and normalized distribution alignment to enable fast adaptation of meta-learners with very few-shot samples, leading to SOTA results on multiple benchmarks in the field of few-shot learning. All of these methods have yielded exceptional results, significantly advancing the field of CDFSL.} These novel perspectives emphasize the technical potential of exploring the adapt strategy and not relying on the CDFSL training strategy. It is worth noting that all these algorithms require access to source data. In contrast, our SF-CDFSL prevents source data access, eliminating privacy and transmission issues. \textcolor{black}{Furthermore, SF-CDFSL provides a robust method and framework for the application of foundational models in CDFSL tasks.}

\subsection{Source-Free Domain Adaptation}
\textcolor{black}{Before SFDA, domain adaptation was widely explored in various fields like remote sensing and medicine. For example, in hyperspectral image classification,~\cite{single1} used a Single-source Domain Expansion Network (SDEnet) to obtain the better performance by employing generative adversarial learning and supervised contrastive learning. Additionally,~\cite{topological1} introduced the Topological structure and Semantic information Transfer network (TSTnet) to align relationships and improve domain adaptation using graph convolutional networks and CNNs.} The first exploration of SFDA~\cite{liang2020we,pei2023uncertainty} suggests that source data is not essential, in which the source classifier parameters will remain frozen to ensure that target representations are aligned with the predictions of source hypotheses. Next, Roy \etal~\cite{roy2022uncertainty} guide target domain adaptation by exploiting uncertainty in source model predictions. Besides, Xia \etal~\cite{xia2021adaptive} incorporate adaptive adversarial inference, contrastive category-wise matching, and self-supervised rotation components to tackle source data unavailability. Moreover, Yang \etal~\cite{yang2021exploiting} notes that although the poor performance of the source model in the target domain is caused by domain gaps, the features generated by the source model tend to cluster. This inspired the authors to utilize the nearest neighbor method to further shorten the distance between similar samples. A growing body of research shows that SFDA often performs better than traditional Unsupervised Domain Adaptation (UDA). However, SFDA source and target tasks are the same. Our SF-CDFSL has domain gaps and task mismatches between the source model and the target task, which makes it more challenging.

\subsection{Transductive Few-Shot Learning}
Certain FSL methods enhance their performance through the strategic use of transductive inference. This entails the integration of query set information during the target task adaptation phase. Notably, TPN~\cite{liu2018learning} introduces and classifies the complete query set to alleviate data scarcity. Simultaneously, TEAM~\cite{qiao2019transductive} fuses meta-learning, metric learning, and transductive inference into a potent cocktail for FSL. Similarly, TRPN~\cite{ma2020transductive} models and propagates the relationships between the support and query sets. Lastly, TIM~\cite{boudiaf2020information} strategy centers on optimizing the mutual information between query features and labels in the FSL tasks. This transductive-based FSL approach consistently performs better than the inductive inference technologies, leveraging the unique ``specific-to-specific'' properties. This aspect is especially advantageous when tackling the more challenging SF-CDFSL problem in comparison to the CDFSL.

\section{Methodology}
\label{m}
In this section, we present a preliminary explanation of the CDFSL and SF-CDFSL problems. We then introduce our proposed approach, Enhanced Information Maximization with Distance-Aware Contrastive Learning (IM-DCL). Finally, we provide an illustration of the adaptation pipeline by IM-DCL.

\subsection{Preliminary}
\subsubsection*{\bf CDFSL}
A CDFSL task consists of a source domain $\mathcal{D}^s$, a source task $\mathcal{T}^s$, as well as a target domain $\mathcal{D}^t$ and a target task $\mathcal{T}^t$ ($\mathcal{D}^s \neq \mathcal{D}^t$ and $\mathcal{T}^s \cap \mathcal{T}^t=\o$). In CDFSL, a large number of supervised data in $\mathcal{D}^s$ is used to process $\mathcal{T}^s$. And $\mathcal{T}^t$ is FSL task, which means there are a few supervised samples in $\mathcal{D}^t$, namely support set $S$. $S=\left\{ x_{i}, y_{i} \right\}^{N \times K}_{i=1}$ contains $N$ classes each of which contains $K$ samples to formulate a $N$-way $K$-shot FSL task, where $x$ and $y$ indicate the training data and their corresponding labels, $x_{i} \in \mathcal{X}_{s}$ and $y_{i} \in \mathcal{Y}_{s}$. While a query set $Q$ is used to evaluate the FSL performance of the model. $Q=\left\{ x_{i} \right\}^{m}_{i=1}$ includes samples from the same $N$ classes with the support set $S$, where $m$ is the sample number of $Q$, $x_{i} \in \mathcal{X}_{q}$. The goal of CDFSL is to obtain a predicting function $f$ for $\mathcal{T}^t$ utilizing both the limited samples in $\mathcal{D}^t$ and the prior knowledge acquired from ($\mathcal{D}^s$, $\mathcal{T}^s$). 

\subsubsection*{\bf SF-CDFSL}
The main difference between SF-CDFSL and CDFSL is accessibility to the source data in $\mathcal{D}^s$. To be more specific, in the SF-CDFSL problem, the source data is inaccessible and its objective is to effectively address the target task $\mathcal{T}^t$, leveraging only the limited supervised samples from $\mathcal{D}^t$ and the source model $\textit{M}$.

\begin{figure*}[!t]
  \centering
  \includegraphics[width=0.8\textwidth]{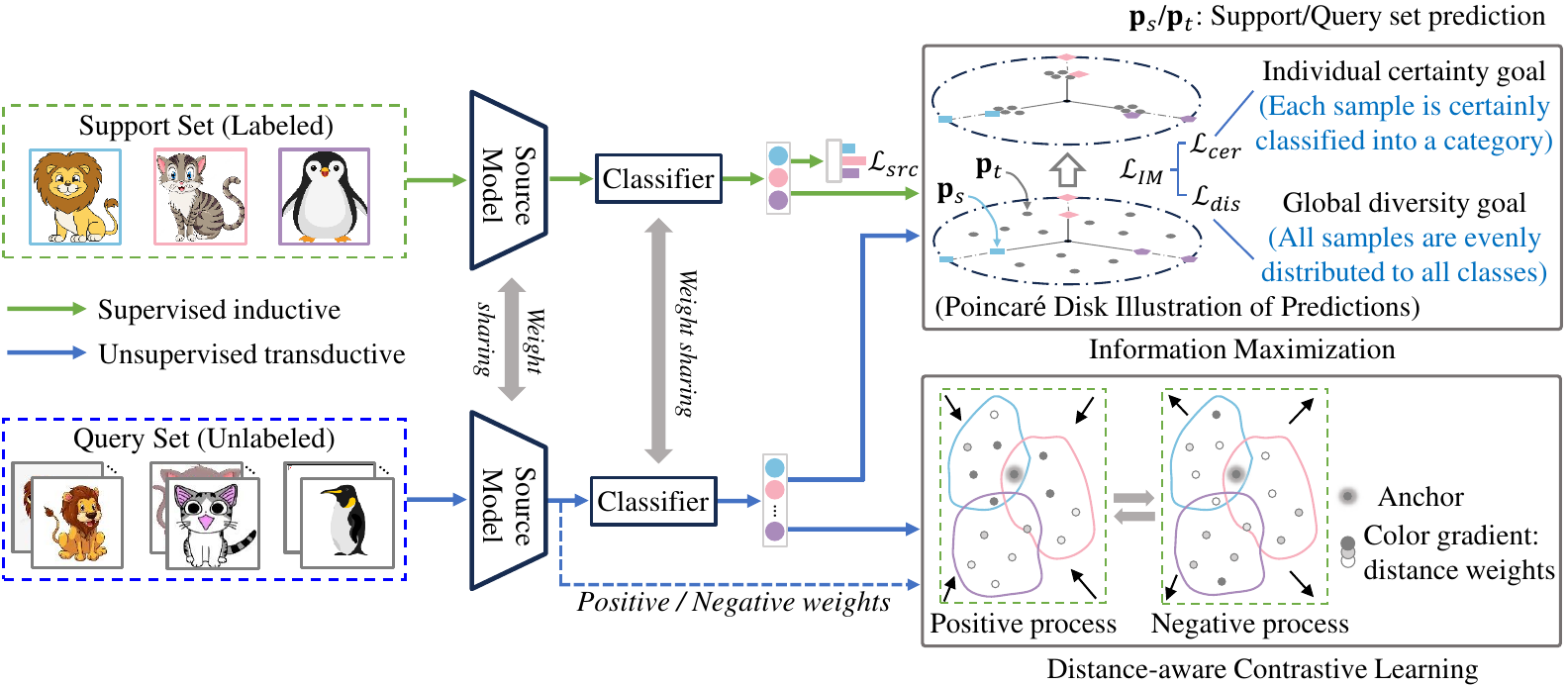}
  \caption{\textcolor{black}{Overview of the proposed IM-DCL method. The images from the support set (with labels) and query set (without labels) are forwarded to the source model to obtain corresponding feature representations. In the supervised inductive process, the support set samples are optimized through supervised CE loss and IM loss. While during transductive process, the proposed DCL was used to enhance IM loss.
  }}
  \label{overview}
  \vspace{-0.3cm}
\end{figure*}

\subsection{IM-DCL Framework}
The source model $\textit{M}$ encodes image samples into a high-dimensional feature space, denoted as $\textit{M}$: $\mathcal{X}_{s} \to \mathbb{R}^{d}$, with $d$ representing the feature dimension. In addition to encoding, $\textit{C}$: $\mathbb{R}^{d} \to \mathbb{R}^{N}$ is typically applied. For solving the SF-CDFSL problem, we introduce an IM-DCL framework, which emphasizes the learning of domain-specific adaptation strategies from the source model $\textit{M}$ and classifier $C$. IM-DCL achieves a dual objective: Firstly, it constrains target samples to generate ideal prediction results through IM. Secondly, DCL is proposed to solve the latent problem of IM failing to learn decision boundaries. Figure~\ref{overview} illustrates the IM-DCL framework overview. The following subsections describe standard supervised fine-tuning, IM transductive inference, DCL transductive inference, and the adaptation pipeline of IM-DCL.

\subsubsection*{\bf Supervised Fine-tuning}
Following the classic FSL procedure, we improve both the source model $\textit{M}$ and the target-specific classifier $\textit{C}$: $f_{\phi}$ = $\textit{C} \circ \textit{M}$: $\mathcal{X}_{s} \to \mathcal{Y}_{s}$, using $S$ in the target domain. Specifically, we input $x_s \in \mathcal{X}_{s}$ into $\textit{M}$ to generate the representations $\textbf{f}_s$ = $\textit{M}$($x_s$), followed by random initialization $\textit{C}$ to predict the outcomes $\textbf{\textit{p}}_s$ = $\textit{C}$($\textbf{f}_s$).
All parameters of $f_{\phi}$ are updated using the following standard supervised cross-entropy (CE) loss,
\begin{equation}
\begin{aligned}
\label{deqn_ex1a}
\mathcal{L}_{src}(f_{\phi};\mathcal{X}_{s},\mathcal{Y}_{s}) =
 -\mathbb{E}_{(x_s,y_s) \in (\mathcal{X}_{s},\mathcal{Y}_{s})}\sum^{N}_{n=1} y_{n} {\rm log} \delta_{n}(\textbf{\textit{p}}_s),
\end{aligned}
\end{equation}
where $\mathbb{E}(\cdot)$ represents the mean value, $\delta_{n}(z)=\frac{exp(z_n)}{\sum_{i}exp(z_i)}$ indicates the $n$-th element in the softmax output of the $N$-dimensional vector $z$. And $y$ is the one-of-$N$ encoding of $y_s$, which means $y_n$ is 1/0 for the correct/wrong class.

\subsubsection*{\bf Information Maximization}
Despite efforts to fine-tune $M$ and $C$ in a supervised manner, the results may not be satisfactory due to domain gaps and a limited amount of labeled target data. Ideally, we would want to optimize $\textit{M}$ and $\textit{C}$ specifically for the target domain, adapting it to the target task. However, since the source data is not accessible, we cannot estimate its distribution. Hence, the core issue we aim to answer is: \textit{What results do we expect when samples are fed into $f_{\phi}$?} Inspired by~\cite{liang2020we}, we argue that each target sample should clearly belong to a category (individual certainty), and all samples should be spread across all categories in a diverse way (global diversity). We achieve this through an information maximization (IM) loss function~\cite{krause2010discriminative,boudiaf2020information,hu2017learning}. On the one hand, to ensure that each target sample is certainly classified into a category, we make each output close to one-hot encoding, which can be achieved by minimizing the entropy of the outputs. The corresponding loss function is as follows:
\begin{equation}
\begin{aligned}
\label{cer}
\mathcal{L}_{cer}(f_{\phi};\mathcal{X}_s,\mathcal{X}_q) = -\mathbb{E}_{x \in (\mathcal{X}_s,\mathcal{X}_q)}\sum^{N}_{n=1} \delta_{n}(f_{\phi}(x)) {\rm log} \delta_{n}(f_{\phi}(x)).
\end{aligned}
\end{equation}
On the other hand, it is crucial to avoid situations where all outputs are similar to the same one-hot encoding, which means we need to maintain diversity of all outputs. We attain this goal through the following equation:
\begin{equation}
\begin{aligned}
\label{div}
& \mathcal{L}_{div}(f_{\phi};\mathcal{X}_s,\mathcal{X}_q) & = \sum^{N}_{n=1} \hat{\textbf{\textit{p}}}_{n} {\rm log} \hat{\textbf{\textit{p}}}_{n},
\end{aligned}
\end{equation}
where $\hat{\textbf{\textit{p}}}_{n}$ indicates the mean embedding of the $n$-th element of all the $N$-dimensional target predictions.
The above $\mathcal{L}_{cer}$ and $\mathcal{L}_{div}$ together constitute the final IM loss,
\begin{equation}
\begin{aligned}
\label{im}
\mathcal{L}_{IM}(f_{\phi};\mathcal{X}_s,\mathcal{X}_q) = \mathcal{L}_{cer}(f_{\phi};\mathcal{X}_s,\mathcal{X}_q) + \lambda_{div} \mathcal{L}_{div}(f_{\phi};\mathcal{X}_s,\mathcal{X}_q),
\end{aligned}
\end{equation}
where $\lambda_{div}$ is the weight of $\mathcal{L}_{div}(f_{\phi};\mathcal{X}_s,\mathcal{X}_q)$. In general, the IM loss $\mathcal{L}_{IM}(f_{\phi};\mathcal{X}_s,\mathcal{X}_q)$ ensures that all outputs are close to one-hot encoding, indicating that they can be confidently classified into a category. Simultaneously, it prevents all outputs from the same one-hot encoding, thus avoiding the situation where all samples are classified into the same class.

\subsubsection*{\bf Distance-aware Contrastive Learning}
\label{mnpc}
\begin{figure}[!t]
  \centering
  \includegraphics[width=0.48\textwidth]{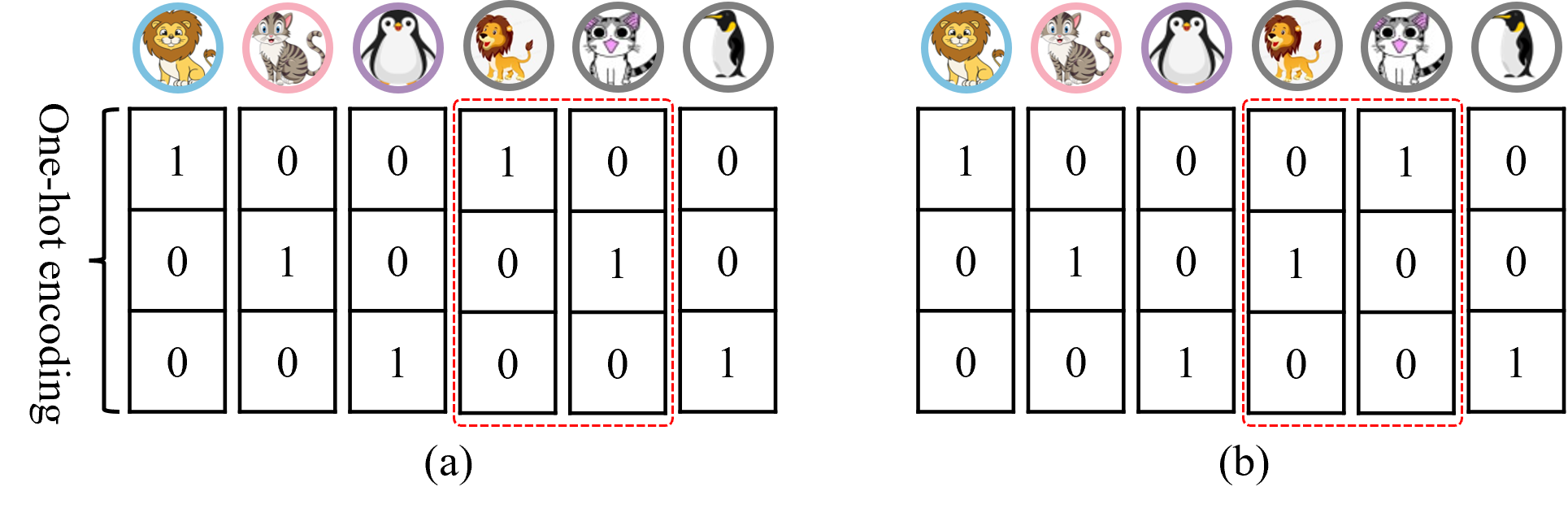}
  \caption{IM cannot tell whether a sample is correctly classified or not. (a) represents the categorical objectives of the samples in the target domain, \ie, the query set is correctly classified. However, problem (b) may arise if IM operates on CDFSL, e.g., if the lion and cat in the query set are classified into each other's category, IM cannot tell this case. Therefore, an DCL algorithm is explored to solve this problem.}
  \label{improblem}
  \vspace{-0.3cm}
\end{figure}

Despite reducing domain disparity between target sample and source model, Information Maximization fails to process target classification. Figure~\ref{improblem} shows that IM cannot avoid distributing samples incorrectly, \ie, IM may classify them into incorrect categories uniformly. This causes a distorted distribution and significantly damages the classification accuracy of the target samples based on the source model.

\begin{figure}[!t]
  \centering
  \includegraphics[width=0.48\textwidth]{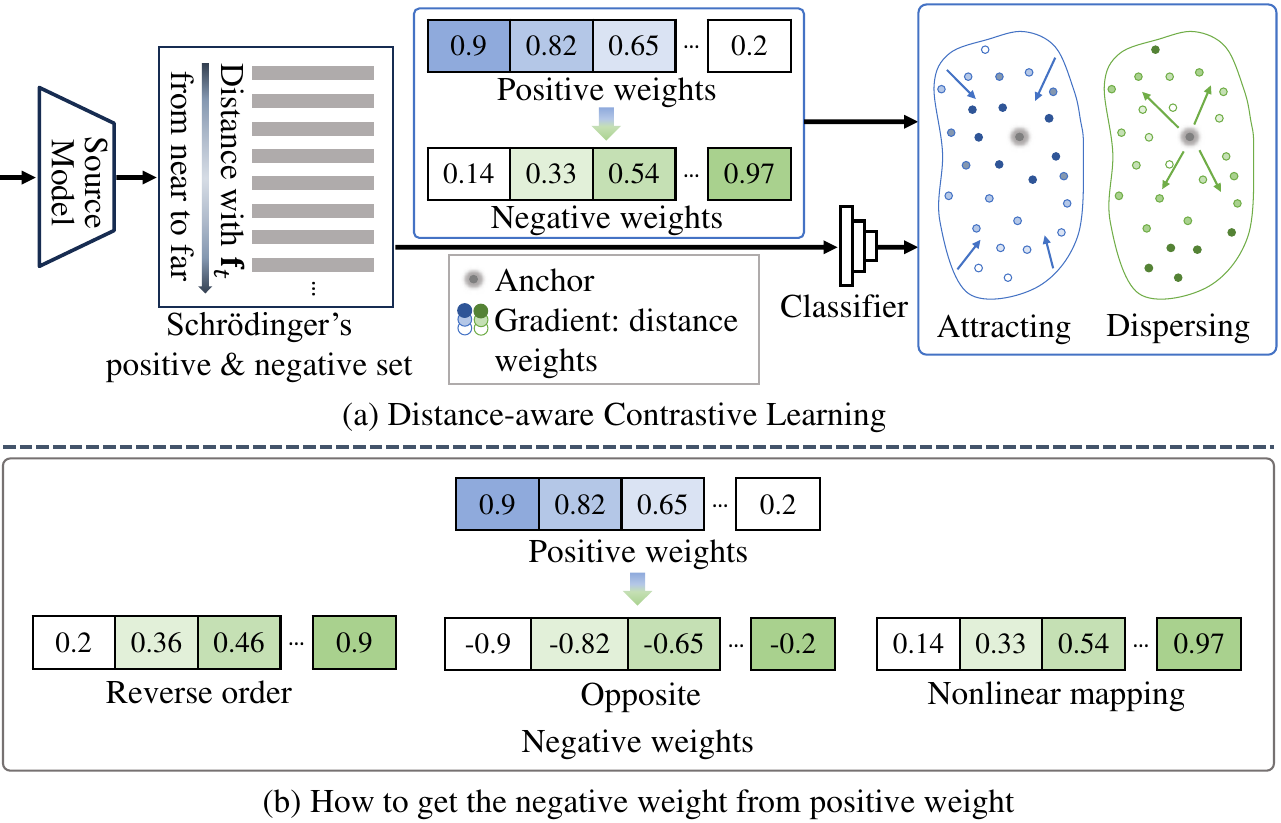}
  \caption{\textcolor{black}{Distance-aware Contrastive Learning algorithm (DCL) achieves contrastive learning by maximizing the similarity of a feature to its positive set while minimizing the similarity to its negative set. (a) illustrates the operation manner of DCL, in which the positive weight matrix are obtained from the distances between all the features and the object feature $\textbf{f}_{t}$. The negative weight matrix is obtained through the positive weight matrix. (b) illustrates the three ways about getting the negative weight matrix from the positive one.} 
  }
  \label{npc}
  \vspace{-0.3cm}
\end{figure}

We identify this issue arising from uncertain outputs near the decision boundary. For instance, the result of the target sample [0.01, 0.49, 0.5, 0, 0] is usually classified into the third class [0, 0, 1, 0, 0], even though it belongs to the second class [0, 1, 0, 0, 0]. As the source model can maintain category consistency in a cross-domain environment, each feature should look more like its neighboring features than the rest~\cite{yang2022attracting}. \textcolor{black}{Taking inspiration from this, this paper attempts to solve it with the contrastive learning (CL) strategy. However, in certain scenarios, categorizing images as strictly positive or negative can prove challenging, especially when these categories lack a clear definition, for instance, in the unlabeled query set. Manual classifying can lead to ambiguous signals, particularly for images situated near critical boundaries~\cite{thoma2020soft}. Furthermore, samples within the same domain exhibit both similarities and differences. As a result, all features inherently contribute both positively and negatively. Therefore,} we propose a Distance-aware Contrastive Learning algorithm (DCL) for further learning decision boundaries in the target domain. Figure~\ref{npc} (a) shows the operation process of DCL. 
\textcolor{black}{Unlike traditional CL algorithms sharply separate positive and negative sets, our proposed DCL introduces a Schrödinger concept where positive and negative sets exhibit shared sample characteristics. Leveraging the same features, DCL crafts distinct weights derived from the distances between all features and the object feature, thus delineating positive and negative sets through weighted feature analysis. Specifically, we first build a memory bank to store all features~\cite{liang2021domain,yang2021generalized}. Then, we select a feature as the object feature $\textbf{f}_{t}$ \textcolor{black}{(also called Anchor, all features in memory bank will be used as anchors to calculate their corresponding positive and negative set weights)}, and the remaining feature set in the memory bank $\{\textbf{f}_i\}^{m-1}_{i=1 }$ as Schrödinger positive and negative set, here, $\textbf{f} = \textit{M}(x)$, where $x \in (\mathcal{X}_s, \mathcal{X}_q)$, $m$ represents the number of all features of the target domain (including support set and query set). Furthermore, we propose a distance-aware approach to softly distinguish positive and negative sets. We first calculate the distance between each feature in $\{\textbf{f}_i\}^{m-1}_{i=1}$ and $\textbf{f}_{t}$,
\begin{equation}
\begin{aligned}
\label{log}
Sim(\textbf{f}_{i},\textbf{f}_{t}) = \frac{\textbf{f}_{i} \cdot \textbf{f}_{t}}{\left \| \textbf{f}_{i} \right \| \left \| \textbf{f}_{t} \right \| } , 
\end{aligned}
\end{equation}
Then we weight each corresponding prediction $\{\textbf{p}_i\}^{m-1}_{i=1}$ with $Sim(\cdot)$ to obtain the real positive set $\mathcal{P}$,
\begin{equation}
\begin{aligned}
\label{log}
\mathcal{P}  = \{ Sim(\textbf{f}_{i},\textbf{f}_{t}) \cdot \textbf{p}_{i} \}^{m-1}_{i=1}, 
\end{aligned}
\end{equation}
Instead, we inversely match the distances to the features and weight them to get the corresponding negative set $\mathcal{N}$,
\begin{equation}
\begin{aligned}
\label{log}
\mathcal{N}  = \{ Sim_{\mathcal{N}}(\textbf{f}_{i},\textbf{f}_{t}) \cdot \textbf{p}_{i} \}^{m-1}_{i=1}, 
\end{aligned}
\end{equation}
where $Sim_{\mathcal{N}}(\cdot)$ means the negative weight, which is inverted through the corresponding positive weight $Sim(\cdot)$. In this paper, we select the three invert manners include reverse order, opposite, and nonlinear mapping, as shown in Figure~\ref{npc} (b). Among them, nonlinear mapping  is implemented through a logical function (variant of the Sigmoid), the formula of which is as follows:
\begin{equation}
\begin{aligned}
\label{log}
Sim_{\mathcal{N}}(\cdot)=\frac{L}{1+e^{(-k\cdot (Sim(\cdot)-x_{0} ))}}, 
\end{aligned}
\end{equation}
where \textit{L} is set to 1, \textit{k} and $x_{0}$ are updated in the finetuning process. Finally, we optimize the following negative log-likelihood formula to implement the contrastive learning operation:}
\begin{equation}
\begin{aligned}
\label{log}
\mathcal{L}_{dcl}(f_{\phi};\mathcal{P},\mathcal{N}) = -{\rm log}\frac{P(\mathcal{P}|f_{\phi})}{P(\mathcal{N}|f_{\phi})},
\end{aligned}
\end{equation}
where both $P(\mathcal{P}|f_{\phi})$ and $P(\mathcal{N}|f_{\phi})$ are two likelihood functions of $\mathcal{P}$ and $\mathcal{N}$:
\textcolor{black}{\begin{equation}
\begin{aligned}
\label{p_log}
P(\mathcal{P}|f_{\phi}) = \prod_{\textbf{p} \in \mathcal{P}} \frac{\textrm{exp}(\textbf{p}_{t}^{\top}\textbf{p})}{\sum^{m}_{k=1}\textrm{exp}(\textbf{p}_{t}^{\top}\textbf{p}_{k})},
\end{aligned}
\end{equation}
\begin{equation}
\begin{aligned}
\label{n_log}
P(\mathcal{N}|f_{\phi}) = \prod_{\textbf{p} \in \mathcal{N}} \frac{\textrm{exp}(\textbf{p}_{t}^{\top}\textbf{p})}{\sum^{m}_{k=1}\textrm{exp}(\textbf{p}_{t}^{\top}\textbf{p}_{k})},
\end{aligned}
\end{equation}
where $\textbf{p}_{t}$ is the prediction of $\textbf{f}_{t}$.} In Eq.~\ref{log}, our objective is to maximize $P(\mathcal{P}|f_{\phi})$ while minimizing $P(\mathcal{N}|f_{\phi})$. However, the calculations about Eq.~\ref{p_log} and Eq.~\ref{n_log} are computationally expensive and resource intensive. Moreover, direct calculation of the likelihood function can result in overfitting. Therefore, instead of optimizing Eq.~\ref{log} directly, we optimize an upper-bound~\cite{yang2022attracting} of Eq.~\ref{log}:
\textcolor{black}{\begin{equation}
\begin{aligned}
\label{sup}
& \mathcal{L}_{dcl}(f_{\phi};\mathcal{P},\mathcal{N}) \\
& \approx -\frac{1}{m} \sum_{t=1}^{m} \sum_{\textbf{p}_{j}\in \mathcal{P}_{t}} \textbf{p}_{t}^{\top}\textbf{p}_{j} +  \frac{\lambda_{\mathcal{N}}}{m} \sum_{t=1}^{m} \sum_{\textbf{p}_{k}\in \mathcal{N}_{t}} \textbf{p}_{t}^{\top}\textbf{p}_{k}, \\
\end{aligned}
\end{equation}
where $\mathcal{P}_{t}$ and $\mathcal{N}_{t}$ are the corresponding positive and negative sets of $\textbf{p}_{t}$. 
This equation has two terms: the first term ensures that the selected target sample $\textbf{p}_{t}$ has high similarity with the positive set, while the second term aims to disperse $\textbf{p}_{t }$ and its corresponding negative set. In this equation, the dot product of two feature represents will be at its peak when both represents are certainly from the same class. $\lambda_{\mathcal{N}}$ reflects the weight assigned to the second term. The weight $\lambda_{\mathcal{N}}$ should gradually decrease as training progresses. It is because the model becomes more precise at later stages of training, allowing it to distinguish target samples from negative samples and to stabilize the distance between them. Hence, it is necessary to weaken the role of dispersing in the second term, as excessive dispersing can push samples from the same class apart.}

\vspace{-0.2cm}
\subsection{Adaptation Pipeline}
In order to make these strategies as effective and as functional as possible, this paper divides the adaptation process into two sub-phases in each epoch, namely the supervised inductive phase and the unsupervised transductive phase. As part of the supervised inductive phase, we optimize the model solely using the support set, using the supervised fine-tuning loss $\mathcal{L}_{src}(f_{\phi};\mathcal{X}_{s},\mathcal{Y}_{s})$ as well as the IM loss $\mathcal{L}_{IM}(f_{\phi};\mathcal{X}_s)$. The objective function for this stage is as follows:
\begin{equation}
\begin{aligned}
\label{ls}
\mathcal{L}_{s}(f_{\phi};\mathcal{X}_{s},\mathcal{Y}_{s}) = \mathcal{L}_{src}(f_{\phi};\mathcal{X}_{s},\mathcal{Y}_{s}) + \lambda_{IM} \mathcal{L}_{IM}(f_{\phi};\mathcal{X}_s),
\end{aligned}
\end{equation}
where $\lambda_{IM}$ is set to 1. In the unsupervised transductive phase, all target samples (including the support and the query set) are processed with enhanced IM which includes IM loss $\mathcal{L}_{IM}(f_{\phi};\mathcal{X}_{s},\mathcal{X}_{q})$ as well as DCL process $\mathcal{L}_{dcl}(f_{\phi};\mathcal{P},\mathcal{N})$. The formula appears as follows:
\begin{equation}
\begin{aligned}
\label{lq}
\mathcal{L}_{q}(f_{\phi};\mathcal{X},\mathcal{P},\mathcal{N}) = \mathcal{L}_{IM}(f_{\phi};\mathcal{X}_{s},\mathcal{X}_{q}) + \lambda_{dcl} \mathcal{L}_{dcl}(f_{\phi};\mathcal{P},\mathcal{N}),
\end{aligned}
\end{equation}
where $\lambda_{dcl}$ = 0.1. 

Algorithm~\ref{pipeline} shows the adaptation pipeline of IM-DCL. At each epoch, the proposed IM-DCL follows the above two-step optimization process for $f_{\phi}$. Initially, in the supervised inductive step, we optimize $f_{\phi}$ by using the support set and $\mathcal{L}_{s}(f_{\phi};\mathcal{X}_{s},\mathcal{Y}_{s})$. This aims to adjust the source model by leveraging the supervised information provided by the support data, ensuring the correct optimization direction. Subsequently, in the unsupervised transductive stage, we further optimize $f_{\phi}$ by utilizing all target data and $\mathcal{L}_{q}(f_{\phi};\mathcal{X},\mathcal{P},\mathcal{N})$. This step enhances the ability of the source model to handle query data by incorporating transductive strategies.
\begin{algorithm}[H]
\footnotesize
\caption{IM-DCL adaptation pipeline.}
\begin{algorithmic}
\STATE
\STATE {\textbf{Input:}} Support set $(\mathcal{X}_s,\mathcal{Y}_s)$, query set $\mathcal{X}_q$, initialized encoder $\textit{M}$ and classifier $\textit{C}$.
\STATE {\textbf{Output:}} Optimized encoder $\textit{M}$ and classifier $\textit{C}$.
\STATE {\textbf{While} \textit{max epochs not reached} \textbf{do}}
\STATE {\hspace{0.5cm} \textbf{for} each mini-batch \{$x_i$,$y_i$\} in $(\mathcal{X}_s,\mathcal{Y}_s)$ \textbf{do}}
\STATE \hspace{1.0cm} $\textbf{f}_s=\textit{M}(\mathcal{X}_s)$
\STATE \hspace{1.0cm} $p_s=\textit{C}(\textbf{f}_s)$
\STATE \hspace{1.0cm} \parbox[t]{\dimexpr0.9\linewidth-\algorithmicindent}{Compute $\mathcal{L}_{s}$ by Eq.~\ref{ls} and update the encoder $\textit{M}$ and classifier $\textit{C}$;}
\STATE \hspace{0.5cm} $\textbf{end for}$
\STATE \hspace{0.5cm} \textbf{for} each {$x_i$} in $(\mathcal{X}_s, \mathcal{X}_q)$ \textbf{do}
\STATE \hspace{1,0cm} $\textbf{f}_t=\textit{M}(x_i)$
\STATE \hspace{1.0cm} \parbox[t]{\dimexpr0.9\linewidth-\algorithmicindent}{\textcolor{black}{Get the positive weight $Sim(\cdot)$ and negative weight $Sim_{\mathcal{N}}(\cdot)$ between the features in memory bank and $\textbf{f}_t$;}}
\STATE \hspace{1.0cm} $\textbf{p}_{t}=\textit{C}(\textbf{f}_t)$
\STATE \hspace{1.0cm} \parbox[t]{\dimexpr0.9\linewidth-\algorithmicindent}{Get the corresponding $\mathcal{P}$ and $\mathcal{N}$ for $\textbf{p}_t$ with $Sim(\cdot)$ and $Sim_{\mathcal{N}}(\cdot)$;}
\STATE \hspace{1.0cm} \parbox[t]{\dimexpr0.9\linewidth-\algorithmicindent}{Compute $\mathcal{L}_{q}$ by Eq.~\ref{lq} and update the encoder $\textit{M}$ and classifier $\textit{C}$;}
\STATE {\hspace{0.5cm} $\textbf{end for}$}
\STATE {\textbf{end while}}
\end{algorithmic}
\label{pipeline}
\end{algorithm}

\section{Experiments and Discussions}
\subsection{Setup}
\subsubsection*{\bf Datasets}
IM-DCL is assessed on 4 datasets from BSCD-FSL~\cite{guo2020broader}, each presenting varying cross-domain challenges: CropDiseases~\cite{mohanty2016using}, EuroSAT~\cite{helber2019eurosat}, ISIC 2018~\cite{tschandl2018ham10000, codella2019skin}, and ChestX~\cite{wang2017chestx}. CropDiseases comprises around 87,000 images of 38 crop leaf types, both healthy and diseased. EuroSAT offers 27,000 labeled images across 10 land use categories with 64x64 resolution. ISIC 2018 contains 2,594 dermoscopic images spanning 7 categories, while ChestX features 112,120 X-ray images from 30,805 patients, annotated for 14 diseases. As per~\cite{guo2020broader}, domain similarity to scene images decreases from CropDiseases to ChestX, covering a spectrum from natural scenes to medical imagery.

\subsubsection*{\bf Baseline Methods}
We conducted extensive evaluations of IM-DCL against multiple CDFSL benchmarks to showcase its effectiveness. Initially, IM-DCL is contrasted with a traditional fine-tuning-based CDFSL benchmark~\cite{guo2020broader}, representing models fine-tuned in the target domain without domain shifts. We then compare IM-DCL to classic meta-learning methods~\cite{vinyals2016matching,finn2017model,sung2018learning,snell2017prototypical,satorras2018few} that use strategies like episodic training. Further, we highlight the robustness of IM-DCL by contrasting it with training strategy-based methods~\cite{fu2023styleadv,wang2021cross,phoo2020self,islam2021dynamic,zheng2023cross} and recent adapted-driven approaches~\cite{yazdanpanah2022visual,oh2022refine,li2022ranking,zhao2023dual,li2023knowledge}. These latter techniques explore domain alignment and feature normalization to enhance CDFSL performance. Notably, all baselines access source data during pre-training. Overall, our broad comparisons underscore the unique strengths and effectiveness of IM-DCL in addressing CDFSL challenges.

\vspace{-0.2cm}
\subsection{Implementation Details}
To maintain objectivity in our evaluations, we use ResNet10 as the model architecture. Initially, we train a model on the \textit{mini}ImageNet~\cite{vinyals2016matching} using standard cross-entropy loss to provide a pre-trained source model (we call CE model). This allows us to evaluate IM-DCL with existing SOTA methods, demonstrating our advantages. Besides, we employ an existing source model, BSR model\footnote{\url{https://github.com/liubingyuu/FTEM\textunderscore BSR\textunderscore CDFSL/tree/master/models/saved/ResNet10\textunderscore bsr\textunderscore aug}}, for further validation. To broaden our perspective on robustness, we also verify the results of IM-DCL on other pre-trained visual architectures like the Vision Transformer (ViT/B-16)\footnote{\url{https://github.com/lucidrains/vit-pytorch}}~\cite{han2022survey}. We use a linear layer for feature classification. Our fine-tuning process employs Stochastic Gradient Descent (SGD). Besides, the learning rate is 0.01, momentum is 0.9, and weight decay is $10^{-3}$. In the IM strategy, we select a $\lambda_{div}$ of 1 to balance prediction certainty and global diversity. Meanwhile, during optimization, the DCL strategy gradually reduces the impact of $\mathcal{N}$ by dynamically adjusting $\lambda_{\mathcal{N}}$. $\lambda_{\mathcal{N}}$ is set to a variable value $(1+10 \times \frac{h_{i}}{H})^{-5.0}$, where $H$ represents the total number of epochs and $h_{i}$ signifies the current epoch. Furthermore, we select 5 features for $\mathcal{P}$, and we set the weight factor $\sigma$ of the support set to 2. Lastly, we set $\lambda_{IM}$ and $\lambda_{dcl}$ to 1 and 0.1, respectively. By illustrating our choices in architecture, optimization settings, and parameters, we ensure experimental repeatability but also a thorough comprehension of the IM-DCL.

\vspace{-0.2cm}
\subsection{Comparison With State-of-the-Art (SOTA) Methods}
We assessed the 5-way 1-shot (5W1S) and 5-way 5-shot (5W5S) tasks on the BSCD-FSL dataset, as shown in Table~\ref{compare}. We begin with traditional CDFSL baselines and classical meta-learning FSL methods, followed by state-of-the-art training strategy-based results. We then showcase adapted strategy-driven CDFSL outcomes and conclude with our proposed IM-DCL. For fairness, IM-DCL is compared against other adapted strategies. We further discuss the results from Table~\ref{compare}.
\begin{table*}
  \caption{The 5W1S and 5W5S results on BSCD-FSL. The proposed IM-DCL clearly achieves SOTA results among all adapted strategy-driven approaches and even performs on par with training strategy-based methods. Bold black values with a deep blue background signal optimal results compared to adaptation-based methods, with a shallow blue background indicating sub-optimal outcomes. And deep and shallow green backgrounds denote the excellent and second excellent results recorded by training strategy-based methods, respectively. \textbf{Source} means access to source data. \textbf{Td} exhibits using a transductive strategy. $\dagger$ represents the method that introduces extra data into the training phase.}
  \label{compare}
  \scriptsize 
  \centering
  \setlength{\tabcolsep}{1.0mm}{
  \begin{tabular}{clcc|ccccc|ccccc}
    \toprule
     \multirow{2}*{\textbf{Type}} & \multirow{2}*{\textbf{Methods}} & \multirow{2}*{\textbf{Source}} & \multirow{2}*{\textbf{Td}} & \multicolumn{5}{c|}{\textbf{5W1S}}  & \multicolumn{5}{c}{\textbf{5W5S}}   \\
    &  &  &  & \textbf{CropDiseases} & \textbf{EuroSAT} & \textbf{ISIC} & \textbf{ChestX} & \textbf{Avg} & \textbf{CropDiseases} & \textbf{EuroSAT} & \textbf{ISIC} & \textbf{ChestX} & \textbf{Avg}   \\
    \midrule
    \multirow{6}*{\rotatebox{90}{Classical}} &  Finetuning~\cite{guo2020broader} & $\checkmark$ & $\times$ & 61.56$\pm$0.90 & 49.34$\pm$0.85 & 30.80$\pm$0.59 &  21.88$\pm$0.38 & 40.90 & 89.25$\pm$0.51 & 79.08$\pm$0.61 & 48.11$\pm$0.64 & 25.97$\pm$0.41   & 60.60   \\
    &  MatchingNet~\cite{vinyals2016matching} & $\checkmark$ & $\times$ & 48.47$\pm$1.01 & 50.67$\pm$0.88 & 29.46$\pm$0.56 & 20.91$\pm$0.30 & 37.38 & 66.39$\pm$0.78 & 64.45$\pm$0.63 & 36.74$\pm$0.53 & 22.40$\pm$0.70  & 47.50 \\
    &  RelationNet~\cite{sung2018learning} & $\checkmark$ & $\times$ & 56.18$\pm$0.85 & 56.28$\pm$0.82 & 29.69$\pm$0.60 & 21.94$\pm$0.42 & 41.02 & 68.99$\pm$0.75 & 61.31$\pm$0.72 & 39.41$\pm$0.58 & 22.96$\pm$0.88  & 48.17 \\
    &  MAML~\cite{finn2017model} & $\checkmark$ & $\times$ & - & - & - & -  & - & 78.05$\pm$0.68 & 71.70$\pm$0.72 & 40.13$\pm$0.58 & 23.48$\pm$0.96  & 53.34 \\

    &  ProtoNet~\cite{snell2017prototypical} & $\checkmark$ & $\times$ & 51.22$\pm$0.50 & 52.93$\pm$0.50 & 29.20$\pm$0.30 & 21.57$\pm$0.20 & 38.73 &  79.72$\pm$0.67 & 73.29$\pm$0.71 & 39.57$\pm$0.57 & 24.05$\pm$1.01  & 54.16 \\
    &  GNN~\cite{satorras2018few} & $\checkmark$ & $\times$ & 64.48$\pm$1.08 & 63.69$\pm$1.03 & 32.02$\pm$0.66 & 22.00$\pm$0.46 & 45.55 & 87.96$\pm$0.67 & 83.64$\pm$0.77 & 43.94$\pm$0.67 & 25.27$\pm$0.46  & 60.20 \\
    \midrule
    \multirow{14}*{\rotatebox{90}{training strategy based}}  &  LRP~\cite{sun2021explanation}      & $\checkmark$ & $\times$ & 59.23$\pm$0.50 & 54.99$\pm$0.50 & 30.94$\pm$0.30 & 22.11$\pm$0.20 & 41.82 & 86.15$\pm$0.40 & 77.14$\pm$0.40 & 44.14$\pm$0.40 & 24.53$\pm$0.30  & 57.99   \\
    &  $\text{FDMixup}^{\dagger}$~\cite{fu2021meta}     & $\checkmark$ & $\times$ & 66.23$\pm$1.03 & 62.97$\pm$1.01 & 32.48$\pm$0.64 & 22.26$\pm$0.45 & 45.99 & 87.27$\pm$0.69 & 80.48$\pm$0.79 & 44.28$\pm$0.66 & 24.52$\pm$0.44   & 59.14   \\
    &  FWT~\cite{tseng2020cross}      & $\checkmark$ & $\times$ & 66.36$\pm$1.04 & 62.36$\pm$1.05 & 31.58$\pm$0.67 & 22.04$\pm$0.44 & 45.59 & 87.11$\pm$0.67 & 83.01$\pm$0.79 & 43.17$\pm$0.70 & 25.18$\pm$0.45  & 59.62   \\
    &  ATA~\cite{wang2021cross}      & $\checkmark$ & $\times$ & 67.47$\pm$0.50 & 61.35$\pm$0.50 & 33.21$\pm$0.40 & 22.10$\pm$0.20 & 46.03 & 90.59$\pm$0.30 & 83.75$\pm$0.40 & 44.91$\pm$0.40 & 24.32$\pm$0.40  & 60.89   \\
    &  AFA~\cite{hu2022adversarial}      & $\checkmark$ & $\times$ & 67.61$\pm$0.50 & 63.12$\pm$0.50 & 33.21$\pm$0.30 & 22.92$\pm$0.20 & 46.72 & 88.06$\pm$0.30 & 85.58$\pm$0.40 & 46.01$\pm$0.40 & 25.02$\pm$0.20  & 61.17   \\
    &  wave-SAN~\cite{fu2022wave} & $\checkmark$ & $\times$ & 70.80$\pm$1.06 & 69.64$\pm$1.09 & 33.35$\pm$0.71 & 22.93$\pm$0.49 & 49.18 & 89.70$\pm$0.64 & 85.22$\pm$0.71 & 44.93$\pm$0.67 & 25.63$\pm$0.49  & 61.37   \\
    &  \text{Confess}~\cite{das2022confess} & $\checkmark$ & $\times$ & - & - & - & -  & - & 88.88$\pm$0.51 & 84.65$\pm$0.38 & 48.85$\pm$0.29 & 27.09$\pm$0.24   &  62.37  \\
    &  $\text{STARTUP}^{\dagger}$~\cite{phoo2020self}     & $\checkmark$ & $\times$ & 75.93$\pm$0.80 & 63.88$\pm$0.84 & 32.66$\pm$0.60 & 23.09$\pm$0.43 & 48.89 & 93.02$\pm$0.45 & 82.29$\pm$0.60 & 47.22$\pm$0.61 & 26.94$\pm$0.44  & 62.37    \\
    &  $\text{ISSNet}^{\dagger}$~\cite{xu2022cross}     & $\checkmark$ & $\times$ & 73.40$\pm$0.86 & 64.50$\pm$0.88 & \cellcolor{green!20}36.06$\pm$0.69 & \cellcolor{green!20}23.23$\pm$0.42 & 49.30 & 94.10$\pm$0.41 & 83.64$\pm$0.55 & 51.82$\pm$0.67 & \cellcolor{green!40}28.79$\pm$0.48  & 64.59  \\
    &  NSAE~\cite{liang2021boosting}  & $\checkmark$ & $\times$ & -          & -          & -          & -         &  - & 93.31$\pm$0.42 & 84.33$\pm$0.55 & \cellcolor{green!40}55.27$\pm$0.62 & 27.30$\pm$0.42  &  65.05   \\
    &  LDP-Net~\cite{zhou2023revisiting} & $\checkmark$ & $\checkmark$ &  69.64 &  65.11 &  33.97 & 23.01 & 47.93 &  91.89 &  84.05 &  48.44 &  26.88 & 65.32 \\
    &  $\text{DDA}^{\dagger}$~\cite{islam2021dynamic} & $\checkmark$ & $\times$ & \cellcolor{green!20}82.14$\pm$0.78 & \cellcolor{green!20}73.14$\pm$0.84 & 34.66$\pm$0.58 & \cellcolor{green!40}23.38$\pm$0.43  & \cellcolor{green!20}53.33 & 95.54$\pm$0.38 & 89.07$\pm$0.47 & 49.36$\pm$0.59 & \cellcolor{green!20}28.31$\pm$0.46  & 65.57    \\
    &  \text{StyleAdv}~\cite{fu2023styleadv}  & $\checkmark$ & $\times$ & 80.69$\pm$0.28 & 72.92$\pm$0.75 & 35.76$\pm$0.52 & 22.64$\pm$0.35 & 53.00 & \cellcolor{green!20}96.51$\pm$0.28 & \cellcolor{green!20}91.64$\pm$0.43 & \cellcolor{green!20}53.05$\pm$0.54 & 26.24$\pm$0.35  & \cellcolor{green!20}66.86    \\
    &  $\text{CLDFD}^{\dagger}$~\cite{zheng2023cross} & $\checkmark$ & $\times$ & \cellcolor{green!40}90.48$\pm$0.72 & \cellcolor{green!40}82.52$\pm$0.76 &   \cellcolor{green!40}39.70$\pm$0.69 & 22.39$\pm$0.44 & \cellcolor{green!40}58.77 & \cellcolor{green!40}96.58$\pm$0.39 & \cellcolor{green!40}92.89$\pm$0.34 & 52.29$\pm$0.62 & 25.98$\pm$0.43  & \cellcolor{green!40}66.94   \\
    \midrule
    \multirow{5}*{\rotatebox{90}{Adapt driven}}  &  KT~\cite{li2023knowledge} & $\checkmark$ & $\checkmark$ & 73.10$\pm$0.87 & 66.43$\pm$0.93 & 34.06$\pm$0.77 & 22.68$\pm$0.60 & 49.07 & 89.53$\pm$0.58 & 82.53$\pm$0.66 & 46.37$\pm$0.77 &  26.79$\pm$0.61  & 61.31    \\
    &  \text{ReFine}~\cite{oh2022refine} & $\checkmark$ & $\times$ & 68.93$\pm$0.84 & 64.14$\pm$0.82 & 35.30$\pm$0.59 & 22.48$\pm$0.41 & 47.64 & 90.75$\pm$0.49 & 82.36$\pm$0.57 &  51.68$\pm$0.63 & 26.76$\pm$0.42  &  62.89   \\
    &  \text{VDB}~\cite{yazdanpanah2022visual}      & $\checkmark$ & $\times$ & 71.98$\pm$0.82 & 63.60$\pm$0.87 & 35.32$\pm$0.65 & \cellcolor{blue!20}22.99$\pm$0.44 & 48.47 & 90.77$\pm$0.49 & 82.06$\pm$0.63 & 48.72$\pm$0.65 & 26.62$\pm$0.45  & 62.04   \\
    &  \text{RDC}~\cite{li2022ranking}  & $\checkmark$ & $\checkmark$ & \cellcolor{blue!40}\textbf{86.33$\pm$0.50} &  \cellcolor{blue!20}71.57$\pm$0.50 &  35.84$\pm$0.40 & 22.27$\pm$0.20 &  \cellcolor{blue!20}54.00 &  93.55$\pm$0.30 &  84.67$\pm$0.30 & 49.06$\pm$0.30 & 25.48$\pm$0.20  &   63.19    \\
    &  DARA~\cite{zhao2023dual} & $\checkmark$ & $\times$ & 80.74$\pm$0.76 & 67.42$\pm$0.80 & \cellcolor{blue!20}36.42$\pm$0.64 & 22.92$\pm$0.40 & 51.88 & \cellcolor{blue!20}95.32$\pm$0.34 & \cellcolor{blue!20}85.84$\pm$0.54 & \cellcolor{blue!40}\textbf{56.28$\pm$0.66} & \cellcolor{blue!20}27.54$\pm$0.42  & \cellcolor{blue!20}66.25    \\
    \midrule
    &  IM-DCL (ours)     & $\times$ & $\checkmark$ &  \cellcolor{blue!20}84.37$\pm$0.99 & \cellcolor{blue!40}\textbf{77.14$\pm$0.71} & \cellcolor{blue!40}\textbf{38.13$\pm$0.57} &  \cellcolor{blue!40}\textbf{23.98$\pm$0.79} & \cellcolor{blue!40}\textbf{55.91} & \cellcolor{blue!40}\textbf{95.73$\pm$0.38} & \cellcolor{blue!40}\textbf{89.47$\pm$0.42} & \cellcolor{blue!20}52.74$\pm$0.69 & \cellcolor{blue!40}\textbf{28.93$\pm$0.41}  &  \cellcolor{blue!40}\textbf{66.72}   \\
    \bottomrule
  \end{tabular}
  }
  \vspace{-0.3cm}
\end{table*}

In the 5W1S task, the proposed IM-DCL shows significant performance advancements over the transfer learning-based benchmark. We achieve remarkable improvements across several datasets, including CropDiseases (increased from 61.56\% to 84.37\%), EuroSAT (enhanced from 49.34\% to 77.14\%), ISIC (boosted from 30.80\% to 38.13\%), and ChestX (raised from 21.88\% to 23.98\%). In general, IM-DCL achieves SOTA performance in the 5W1S tasks, recording an average accuracy of 55.91\% and surpassing all existing adapted strategy-driven methods. In particular, our approach achieves SOTA results on EuroSAT, ISIC and ChestX, registering 77.14\%, 38.13\%, and 23.98\%, respectively. \textcolor{black}{Compared to the 84.37\% performance of IM-DCL on CropDisease, RDC~\cite{li2022ranking} achieved a higher result of 86.33\%. This enhancement is attributed to RDC's integration of a distance matrix, transformed through a Hyperbolic tangent transformation, with the Normalised non-linear subspace's distance matrix, effectively preserving more transferable discriminative information. This is particularly beneficial for the CropDisease task, a cross-near domain challenge, which inherently contains a higher volume of transferable data, leading to improved performance in this near-domain scenario. Moreover, in contrast to training strategy-based approaches, IM-DCL also displayed competitive effectiveness. Compared to the SOTA CLDFD~\cite{zheng2023cross} method, which achieved 90.48\%, 82.52\%, and 39.70\% on the CropDisease, EuroSAT, and ISIC datasets respectively, IM-DCL performed slightly less impressively. A notable observation is CLDFD's enhanced performance in near-domains, partly due to the incorporation of target domain information during the training phase, thereby endowing the model with a certain level of generalizability to target domain samples from the outset. As IM-DCL was specifically developed for SF-CDFSL, it lacks the provision for any additional modifications during the pre-training phase. Besides, CLDFD has achieved SOTA results in terms of average performance, while IM-DCL has also reached SOTA outcomes when compared with adaptation-driven algorithms. Additionally, the other SOTA methods~\cite{islam2021dynamic,yazdanpanah2022visual,fu2023styleadv,zhao2023dual} have also achieved highly competitive performance levels.}

Likewise, IM-DCL exhibits good performance over the baseline in the 5W5S task, delivering an average enhancement of 6.04\%. Compared with adapted strategy-driven methods, IM-DCL achieves SOTA overall average performance with average accuracy of 66.72\%. Moreover, it achieves stellar outcomes on different datasets, \ie, 95.73\% for CropDiseases, 89.47\% for EuroSAT, 52.44\% for ISIC, and 28.93\% for ChestX. \textcolor{black}{However, compared to DARA's performance on four datasets, with 95.32\%, 85.84\%, 56.28\%, and 27.54\% respectively, IM-DCL experienced a 3.84\% decline in performance on the ISIC dataset. This drop may be attributed to the distinctive nature of the ISIC dataset, where, unlike the other three datasets featuring images with clean backgrounds, ISIC images are susceptible to background interference. For instance, images in the 'melanocytic nevus' category containing hair might lead the model to mistakenly identify hair as a pathological feature.} It is worth noting that IM-DCL generates competitive results compared to training strategy-based methods. \textcolor{black}{CLDFD achieved SOTA results in near-domain tasks, with 96.58\% on CropDisease and 92.89\% on EuroSAT. Furthermore, NSAE also demonstrated competitive performance on the ISIC dataset. And StyleAdv~\cite{fu2023styleadv} obtains the sub-optimal results on CropDiseases (96.51\%), EuroSAT (91.64\%), and ISIC (53.05\%). Meanwhile, on ChestX, the proposed IM-DCL achieved a SOTA result of 28.93\%. Finally, we observe that even though training strategy-based methods incorporate various strategies during the pre-training phase, their performance in distant domains is not overwhelmingly superior. For instance, the state-of-the-art (SOTA) methods based on training strategies achieved 55.27\% and 28.79\% on the ISIC and ChestX datasets, respectively, while the adapted strategy-driven SOTA methods attained similar results with 56.28\% and 28.93\% in these domains. This suggests that, compared to near-domain tasks, adapted strategy-driven methods often demonstrate enhanced capabilities in handling distant domain tasks.}

\begin{figure*}[!t]
  \centering
  \includegraphics[width=0.9\textwidth]{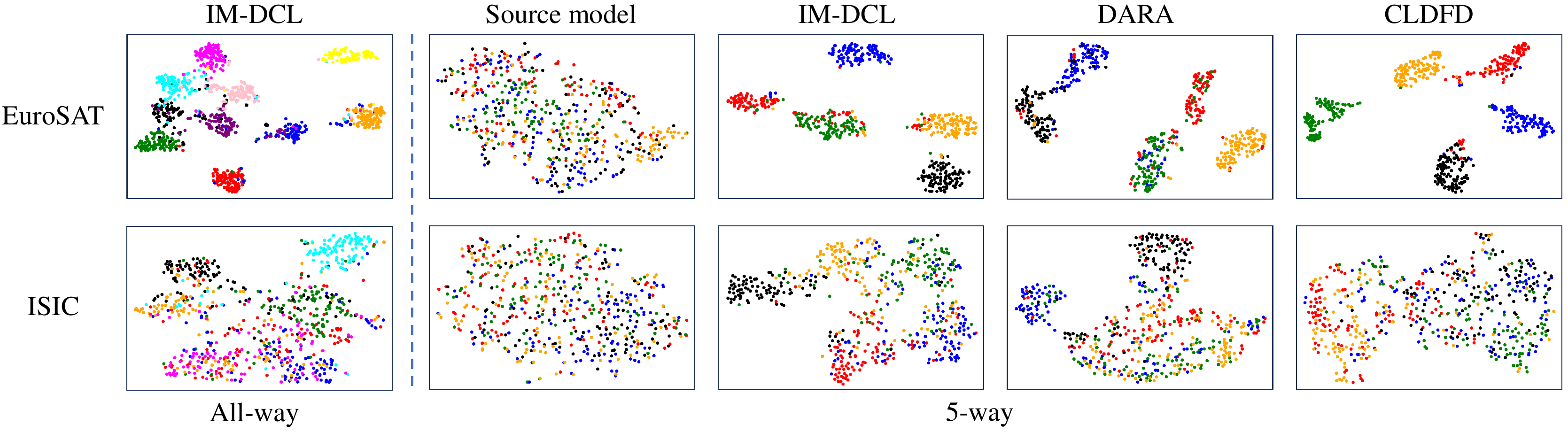}
  \caption{\textcolor{black}{Visualization results on EuroSAT and ISIC. Left part means the results on all-way task. Right part shows the visualization on 5-way task. Different color means different categories. Source model means that the result when the source model is not fine-tuned.}}
  \label{imnpcfc}
  \vspace{-0.3cm}
\end{figure*}

\textcolor{black}{Furthermore, to elucidate the strengths and limitations of IM-DCL in comparison to SOTA methods, we present tSNE visualizations for both all-way and 5-way tasks conducted on the EuroSAT and ISIC datasets. These visualizations are showcased in Figure~\ref{imnpcfc}. Our choice of these datasets is rooted in their unique characteristics.} EuroSAT, a remote sensing RGB image dataset, has relatively small domain gaps with most pre-trained source models trained on extensive natural scene datasets. Therefore, the visualization results of IM-DCL on this dataset represent the performance in near domain scenarios. Conversely, as a medical dataset, ISIC is significantly distanced from the source model in terms of both the domain and task, reflecting the performance of IM-DCL in distant domain scenarios. It is worth mentioning that ChestX, a dataset further distanced from the source model, was not chosen for visualization. The main reason is, due to the substantial domain gap, none of the existing SOTA methods have made substantial breakthroughs on ChestX (with a typical 25\%-28\% performance on 5W5S). Minor performance improvements are challenging to distinguish in visualizations.

\textcolor{black}{Firstly, the proposed IM-DCL demonstrates commendable visualization results for both the near-domain (EuroSAT) and distant-domain (ISIC) scenarios in the all-way task, as depicted in the Figure~\ref{imnpcfc}. However, in the standard 5-way task, CLDFD~\cite{zheng2023cross} notably outperforms other methods on the EuroSAT dataset. For ISIC, the subtleties in visualizations may not be immediately apparent, primarily due to the prevailing challenge of low performance across methods in distant-domain tasks. Nevertheless, t-SNE analysis unveils the confidence level is elevated with IM-DCL and DARA~\cite{zhao2023dual} on the ISIC dataset. These insights emphasize the effective of CLDFD on EuroSAT. While ISIC presents inherent difficulties, the t-SNE visualizations provide valuable insights into DARA's and IM-DCL's promising results within this context.}

\textcolor{black}{In general, compared to existing SOTA methods, the proposed approach offers the following three advantages: 1) IM-DCL achieves SOTA performance under the source-free scenes, meaning it addresses the FSL tasks in the target domain without accessing to source domain data, which resolves potential data privacy and transfer issues arising from accessing source data. 2) IM-DCL is model-free, which means it works without any pretraining strategy about the model, indicating that the proposed method is applicable to any classification model. 3) The proposed IM-DCL can advance the application of existing base models in CDFSL. 4) IM-DCL achieves SOTA performance in the target domain, particularly showing significant potential for performance improvement in distant domain FSL tasks. Research based on this has the potential to break through the performance limitations of existing methods in distant domains. However, due to the unique challenges of SF-CDFSL, IM-DCL also has some potential downsides: 1) It relies on existing pre-trained models to address FSL tasks in the target domain, and for a given pre-trained model, it cannot introduce target domain information during the pre-training process to make the model more target-domain-friendly, especially in distant domains. 2) Compared to other CDFSL methods, only a small amount of labeled target domain data is available. The proposed method needs to achieve desirable results using only a limited number of samples, which is quite challenging. 3) Results in data with complex background interference, such as ISIC, need to be improved.}

\vspace{-0.2cm}
\subsection{Ablation Study}
IM-DCL comprises two primary components: Information Maximization (IM) and Distance-aware Contrastive Learning (DCL). This section evaluates their contributions against a transfer learning baseline, as shown in Table~\ref{imnpct}. For the role of IM in IM-DCL, we assess its performance in both non-transductive (by adding IM to the support set) and transductive contexts (integrating IM across all target samples). We also examine the impacts of individual certainty and global diversity on IM. Furthermore, DCL performance is first validated by comparing its outcomes with the traditional contrastive learning.
\textcolor{black}{Afterward, we compare the performance of the proposed DCL and the existing contrastive strategy InfoNCE~\cite{infonce}, as shown in Table~\ref{InfoNCE}, in which we evaluate the effects of InfoNCE $\textit{w/}$ or $\textit{w/o}$ negative samples.} 
Lastly, we assess the effects of varying $\lambda_{\mathcal{N}}$ values in $\mathcal{N}$.

\begin{table}
  \caption{Performance comparison of different components. First, the effects of IM and DCL are evaluated, then the effects of the transductive and non-transductive introduction of IM are verified. In addition, the effect of weight matrix is shown by comparing `IM+DCL $w/o$' and `IM-DCL'. `IM+DCL $w/o$' means DCL without the weight matrix. `$K$' means 5-way $K$-shot task.}
  \label{imnpct}
  \scriptsize 
  \centering
  \setlength{\tabcolsep}{1.0mm}{
  \begin{tabular}{c|lccccc}
    \toprule
    \textbf{\textit{K}} &  \textbf{Methods} & \textbf{CropDiseases} & \textbf{EuroSAT} & \textbf{ISIC} & \textbf{ChestX} & \textbf{Avg}   \\
    \midrule
    \multirow{7}*{1} & Fine-tuning  & 61.56$\pm$0.90 & 49.34$\pm$0.85 & 30.80$\pm$0.59 &  21.88$\pm$0.38 & 40.90   \\
    & SIM & 73.07$\pm$0.83 & 63.82$\pm$0.81 & 35.65$\pm$0.64 & 22.72$\pm$0.40 & 48.82    \\
    &  IM  & 82.59$\pm$0.91 & 73.12$\pm$0.89 & 38.65$\pm$0.74 & 22.28$\pm$0.43 & 54.16    \\
    & IM+DCL $w/o$ & 82.85$\pm$0.90 & 73.45$\pm$1.02 & 39.01$\pm$0.75 & 22.80$\pm$0.43 & 54.53    \\
    & SIM+DCL $w/o$  & 83.06$\pm$0.90 & 74.71$\pm$0.89 & 38.37$\pm$0.76 & 22.50$\pm$0.43 & 54.66    \\
    & IM+DCL $w/o$  & 83.89$\pm$0.94 & 74.25$\pm$0.96 & \textbf{39.32$\pm$0.76} & 22.67$\pm$0.43 & 55.03    \\
    & IM-DCL  & \textbf{84.37$\pm$0.99} & \textbf{77.14$\pm$0.71} & 38.13$\pm$0.57 & \textbf{23.98$\pm$0.79} & \textbf{55.91}   \\
    \midrule
    \multirow{7}*{5} & Fine-tuning  & 89.25$\pm$0.51 & 79.08$\pm$0.61 & 48.11$\pm$0.64 & 25.97$\pm$0.41 & 60.60   \\
    & SIM  & 91.95$\pm$0.47 & 82.88$\pm$0.59 & 52.53$\pm$0.63 & 27.57$\pm$0.44 & 63.73    \\
    & IM  & 94.35$\pm$0.42 & 86.02$\pm$0.52 & 53.69$\pm$0.67 & 27.90$\pm$0.43 & 65.49    \\
    & IM+DCL $w/o$  & 95.11$\pm$0.40 & 87.60$\pm$0.53 & 54.72$\pm$0.70 & 28.18$\pm$0.45 & 66.40    \\
    & SIM+DCL $w/o$ & 94.99$\pm$0.39 & 88.32$\pm$0.50 & 54.31$\pm$0.74 & 27.96$\pm$0.46 & 66.40    \\
    & IM+DCL $w/o$  & 94.86$\pm$0.41 & 87.92$\pm$0.52 & \textbf{55.21$\pm$0.66} & 28.15$\pm$0.48 & 66.54    \\
    & IM-DCL  &  \textbf{95.73$\pm$0.38} & \textbf{89.47$\pm$0.42} & 52.74$\pm$0.69 & \textbf{28.93$\pm$0.41}  &  \textbf{66.72}   \\
    \bottomrule
  \end{tabular}
  }
  \vspace{-0.3cm}
\end{table}

\begin{figure*}[!t]
  \centering
  \includegraphics[width=0.9\textwidth]{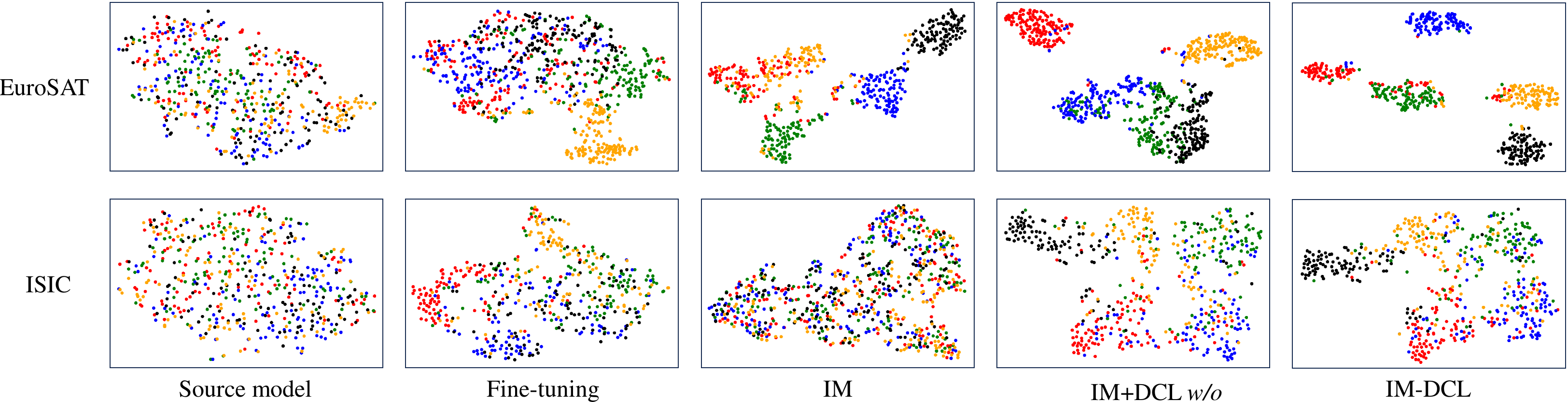}
  \caption{\textcolor{black}{Visualization results of the 5W1S task. Different color means different categories. Source model means that the result when the source model is not fine-tuned. Compared to the result of Source model, Fine-tuning help to classify.}}
  \label{ablationv}
  \vspace{-0.3cm}
\end{figure*}

\textcolor{black}{In addition, to illustrate the impact of each component intuitively, we present t-SNE visualization results for the 5W5S task on the EuroSAT and ISIC datasets, as displayed in Figure~\ref{ablationv}. Specifically, the figure shows the effectiveness of IM-DCL in the near and far domains by comparing the tSNE visualization results of Source model, Fine-tuning, IM, IM+DCL $w/o$, and IM-DCL.}

\subsubsection*{\bf Effect of Information Maximization (IM)}
First, we evaluate the influence of information maximization (IM) through transductive and non-transductive usage. The results, displayed in Table~\ref{imnpct}, cover several scenarios: `SIM' points to the application of IM only in the support set, embodying non-transductive IM usage. `IM' implies the integration of IM across all target data, marking transductive IM usage. `DCL $w/o$' signifies ignoring the weights in DCL. `IM+DCL $w/o$' fuses both `IM' and `DCL $w/o$'. Lastly, ` IM-DCL' highlights the concurrent usage of both IM and weighted DCL.

The comparison of `Fine-tuning', `SIM', and `IM' facilitates the evaluation of IM efficacy. We can know from the table that `IM' excels in both 5W1S and 5W5S tasks, reaching average accuracy of 54.16\% and 65.49\%, respectively. In the 5W1S task, `IM' surpasses `SIM' (with an overall accuracy of 48.82\%) by an impressive 5.34\%. Additionally, `IM' secures superior results on 3 datasets (excluding ChestX), expressing higher scores (82.59\%, 73.12\%, and 38.65\%) compared to `Fine-tuning' and `SIM'. In the 5W5S task, `IM' outperforms the former two across all four datasets. Furthermore, `SIM' outperforms `Fine-tuning' in both 5W1S and 5W5S tasks, verifying the substantial improvement of IM, even when employed exclusively to the support set. Additionally, the assessment of `DCL $w/o$' and `IM+DCL $w/o$' substantiates the positive influence of IM when intersected with DCL. The findings invariably demonstrate the advantage of `IM+DCL $w/o$'. Therefore, the significant enhancement of IM in CDFSL is unquestionable. Moreover, Figure~\ref{ablationv} visually express the impact of IM by contrasting `Fine-tuning' and `IM' and clearly exhibit the improvements of IM. Nonetheless, owing to the large domain divergence between ISIC and the source model, the visual enhancement attributable to `IM' is not as conspicuous, which corresponds with the numerical enhancement degree observed in Table~\ref{imnpct}.

In addition, we compare the influence of individual certainty (IC) and global diversity (OD) in IM respectively. Table~\ref{imt} presents the results of keeping only one of the two in IM. Notably, the elimination of IC (`IM \textit{w/o IC}') results in the lowest averages of 53.35\% and 59.05\% for the 5W1S and 5W5S tasks, respectively. These scores represent a significant drop of 0.81\% and 6.44\% compared to the IM results (54.16\% and 65.49\%). It means `IM \textit{w/o IC}' exhibits a substantial gap in performance on all 4 datasets compared to IM, underscoring the crucial role of IC in IM. However, the absence of OD (`IM \textit{w/o OD}') produces a less dramatic effect, with averages of 53.81\% and 65.13\% on the 5W1S and 5W5S tasks, respectively. The improvement due to OD in IM is less pronounced, amounting to only 0.35 and 0.36. Interestingly, `IM \textit{w/o OD}' actually surpasses IM performance on the EuroSAT dataset in the 5W5S task, scoring 86.04\% compared to 86.02\% in IM. In summary, both IC and OD help IM improve the CDFSL performance, while IC plays a pivotal role in IM, and OD weakly enhances IM performance relatively.
\begin{table}
  \caption{The role of individual certainty (IC) and global diversity (OD) of IM in the 5-way $K$-shot tasks.}
  \label{imt}
  \scriptsize 
  \centering
  \setlength{\tabcolsep}{1.0mm}{
  \begin{tabular}{l|lccccc}
    \toprule
    \textbf{\textit{K}} &  \textbf{Methods} & \textbf{CropDiseases} & \textbf{EuroSAT} & \textbf{ISIC} & \textbf{ChestX} & \textbf{Avg}   \\
    \midrule
    \multirow{3}*{1} & IM \textit{w/o IC} & 80.19$\pm$0.63 & 72.34$\pm$0.95 & 37.93$\pm$0.26 & 22.93$\pm$0.73 & 53.35    \\
     & IM \textit{w/o OD}  & 81.83$\pm$0.83 & 73.04$\pm$0.89 & 38.14$\pm$0.77 & 22.22$\pm$0.44 & 53.81   \\
     & IM & \textbf{82.59$\pm$0.91} & \textbf{73.12$\pm$0.89} & \textbf{38.65$\pm$0.74} & \textbf{22.28$\pm$0.43} & \textbf{54.16}    \\
    \midrule
    \multirow{3}*{5} & IM \textit{w/o IC} & 93.25$\pm$0.41 & 72.35$\pm$0.70 & 45.49$\pm$0.60 & 25.10$\pm$0.40 & 59.05    \\
    & IM \textit{w/o OD}  & 94.24$\pm$0.39 & \textbf{86.04$\pm$0.51} & 53.34$\pm$0.67 & 26.88$\pm$0.47 & 65.13   \\
    & IM  & \textbf{94.35$\pm$0.42} & 86.02$\pm$0.52 & \textbf{53.69$\pm$0.67} & \textbf{27.90$\pm$0.43} & \textbf{65.49}    \\
    \bottomrule
  \end{tabular}
  }
  \vspace{-0.3cm}
\end{table}

\subsubsection*{\bf Effect of Distance-aware Contrastive Learning (DCL)}
To evaluate the contribution of the DCL, we analyze the performance of `SIM' and `SIM+DCL $w/o$', which underlines the enhancement delivered by the application of `DCL $w/o$'. As shown in Table~\ref{imnpct}, `SIM+DCL $w/o$' outshines `SIM' by a significant margin of 5.84\% in the 5W1S task and 2.67\% in the 5W5S task. The comparison between `IM' and `IM+DCL $w/o$' further illuminates the efficacy of `DCL $w/o$'. In the 5W5S task, `IM+DCL $w/o$' attains better results than `IM' across all datasets. 
Moreover, the proposed ` IM-DCL' performs 0.18\% better than `IM+DCL $w/o$' on average and surpasses it on 3 datasets (except ISIC), thereby endorsing the benefits of weighted information into DCL. The inferior performance of ` IM-DCL' on ISIC compared to `IM+DCL $w/o$' is attributed to the minimal color gradient between the foreground and the background of images in ISIC. This implies that the classification is influenced by background elements that are irrelevant to the main information. For example, images with hair in the `melanocytic nevus' category may lead the model to establish hair as a distinguishing feature of pathology. The introduction of positive and negative weights in IM-DCL exacerbates this phenomenon, proving that weights are not amenable to samples with indistinct color gradients. From Figure~\ref{ablationv}, we visualize the impact of `DCL $w/o$' through the comparison between `IM' and `IM+DCL $w/o$'. Further verification of the effectiveness of positive and negative weights come from comparing `IM+DCL $w/o$' with ` IM-DCL' in Figure~\ref{ablationv}.

\textcolor{black}{In addition to the traditional contrastive learning strategy, we first weight the samples of the positive set by distance based on traditional contrastive learning, and then perform contrastive learning calculations. The results are displayed in IM+DCL $w/$ in the Table~\ref{InfoNCE}. we also compare the performance of the proposed DCL with the existing InfoNCE, as shown in Figure~\ref{InfoNCE}. We compared two different usages of InfoNCE, \ie, InfoNCE $\textit{with/without}$ negative sets. From the Table, it can be seen that the proposed IM-DCL achieved the highest average performance of 55.91\% and 66.72\% on 5W1S and 5W5S respectively. However, based on the same reasons as traditional contrastive learning, InfoNCE also achieved better results on ISIC than the proposed DCL. For example, on the 5W5S task, compared to DCL 52.44\% of the results, InfoNCE achieved better 55.42\% results on ISIC. In addition, the performance comparison with IM+DCL $w/$ also shows that besides ISIC, the proposed IM-DCL achieves higher performance on both 5W1S and 5W5S tasks on the other data sets.}
\begin{table}
  \caption{\textcolor{black}{Performance of different contrastive learning strategy. DCL $w/o$ indicates the traditional contrastive learning strategy, which splits the positive and negative sets. DCL $w/$ means using distances to weight the positive set. InfoNCE $\textit{w/}$ means InfoNCE with Negative samples, and InfoNCE $\textit{w/o}$ means InfoNCE without Negative sample.}}
  \label{InfoNCE}
  \scriptsize 
  \centering
  \setlength{\tabcolsep}{0.8mm}{
  \color{black}
  \begin{tabular}{c|lccccc}
    \toprule
    \textbf{\textit{K}} &  \textbf{Methods} & \textbf{CropDiseases} & \textbf{EuroSAT} & \textbf{ISIC} & \textbf{ChestX} & \textbf{Avg}   \\
    \midrule
    \multirow{7}*{1} & Fine-tuning  & 61.56$\pm$0.90 & 49.34$\pm$0.85 & 30.80$\pm$0.59 &  21.88$\pm$0.38 & 40.90   \\
    & IM+DCL $w/o$  & 83.89$\pm$0.94 & 74.25$\pm$0.96 & \textbf{39.32$\pm$0.76} & 22.67$\pm$0.43 & 55.03    \\
    & IM+DCL $w/$  & 84.10$\pm$0.90 & 74.42$\pm$0.92 & 38.93$\pm$0.74 & 22.98$\pm$0.42 & 55.11   \\
    & \textcolor{black}{IM+InfoNCE \textit{w/o}}  & \textcolor{black}{\textbf{85.69$\pm$0.63}} & \textcolor{black}{73.78$\pm$0.53} & \textcolor{black}{37.91$\pm$0.31} & \textcolor{black}{23.31$\pm$0.88} & \textcolor{black}{55.17}    \\
    & \textcolor{black}{IM+InfoNCE \textit{w/}}  & \textcolor{black}{84.58$\pm$0.46} & \textcolor{black}{76.27$\pm$0.54} & \textcolor{black}{37.80$\pm$0.73} & \textcolor{black}{21.78$\pm$0.95} & \textcolor{black}{55.11}    \\
    & IM-DCL  & 84.37$\pm$0.99 & \textbf{77.14$\pm$0.71} & 38.13$\pm$0.57 & \textbf{23.98$\pm$0.79} & \textbf{55.91}   \\
    \midrule
    \multirow{7}*{5} & Fine-tuning  & 89.25$\pm$0.51 & 79.08$\pm$0.61 & 48.11$\pm$0.64 & 25.97$\pm$0.41 & 60.60   \\
    & IM+DCL $w/o$  & 94.86$\pm$0.41 & 87.92$\pm$0.52 & \textbf{55.21$\pm$0.66} & 28.15$\pm$0.48 & 66.54    \\
    & IM+DCL $w/$ & 95.18$\pm$0.40 & 88.16$\pm$0.51 & 54.93$\pm$0.92 & 28.41$\pm$0.47 & 66.67   \\
    & \textcolor{black}{IM+InfoNCE \textit{w/o}} & \textcolor{black}{95.47$\pm$0.89} & \textcolor{black}{87.78$\pm$0.21} & \textcolor{black}{54.42$\pm$0.92} & \textcolor{black}{28.53$\pm$0.52} & \textcolor{black}{66.55}    \\
    & \textcolor{black}{IM+InfoNCE \textit{w/}}  & \textcolor{black}{93.33$\pm$0.28} & \textcolor{black}{87.27$\pm$0.83} & \textcolor{black}{54.36$\pm$0.59} & \textcolor{black}{28.40$\pm$0.51} & \textcolor{black}{65.84}    \\
    & IM-DCL  &  \textbf{95.73$\pm$0.38} & \textbf{89.47$\pm$0.42} & 52.74$\pm$0.69 & \textbf{28.93$\pm$0.41}  &  \textbf{66.72}   \\
    \bottomrule
  \end{tabular}
  }
  \vspace{-0.3cm}
\end{table}

\begin{figure}[!t]
  \centering
  \includegraphics[width=0.45\textwidth]{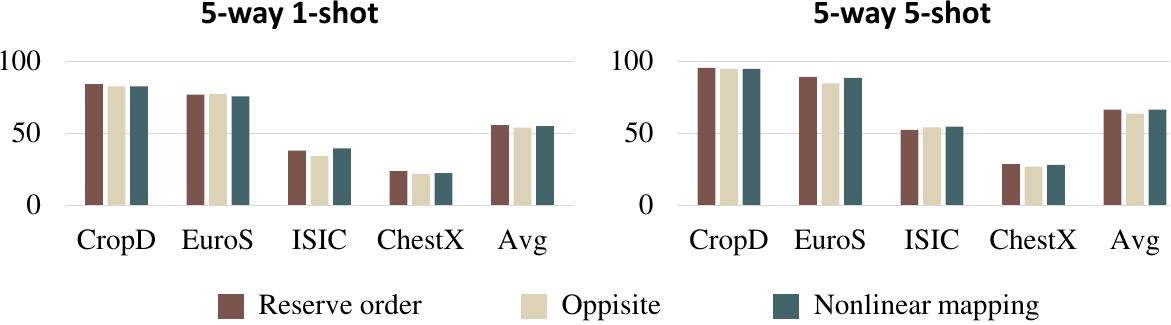}
  \caption{Results of different negative weight generate manner. `CropD' means the dataset CropDiseases, `EuroS' indicates EuroSAT.}
  \label{manner1}
\end{figure}

\textcolor{black}{Finally, we evaluate the three generation manners of negative weight. As shown in Figure~\ref{manner1}, the reverse order way obtains the optimal average result on 5W1S. However, reserve ordered negative weight get the sub-optimal average result on 5W5S task. The main reason is because compared to the other two manners obtain the better results on ISIC, reserve ordered negative weight.}

\textcolor{black}{
\subsubsection*{\bf Effect of Loss Hyperparameters}
The hyperparameters, such as $\lambda_{div}$, $\lambda_{IM}$, $\lambda_{\mathcal{N}}$ and $\lambda_{dcl}$, of loss functions in this paper are set as 1, 1, variable, and 0.1, respectively. Figure~\ref{parameter_select} indicates the results of different values of $\lambda_{div}$, $\lambda_{IM}$, and $\lambda_{dcl}$.
\begin{figure*}
  \centering
  \includegraphics[width=0.9\textwidth]{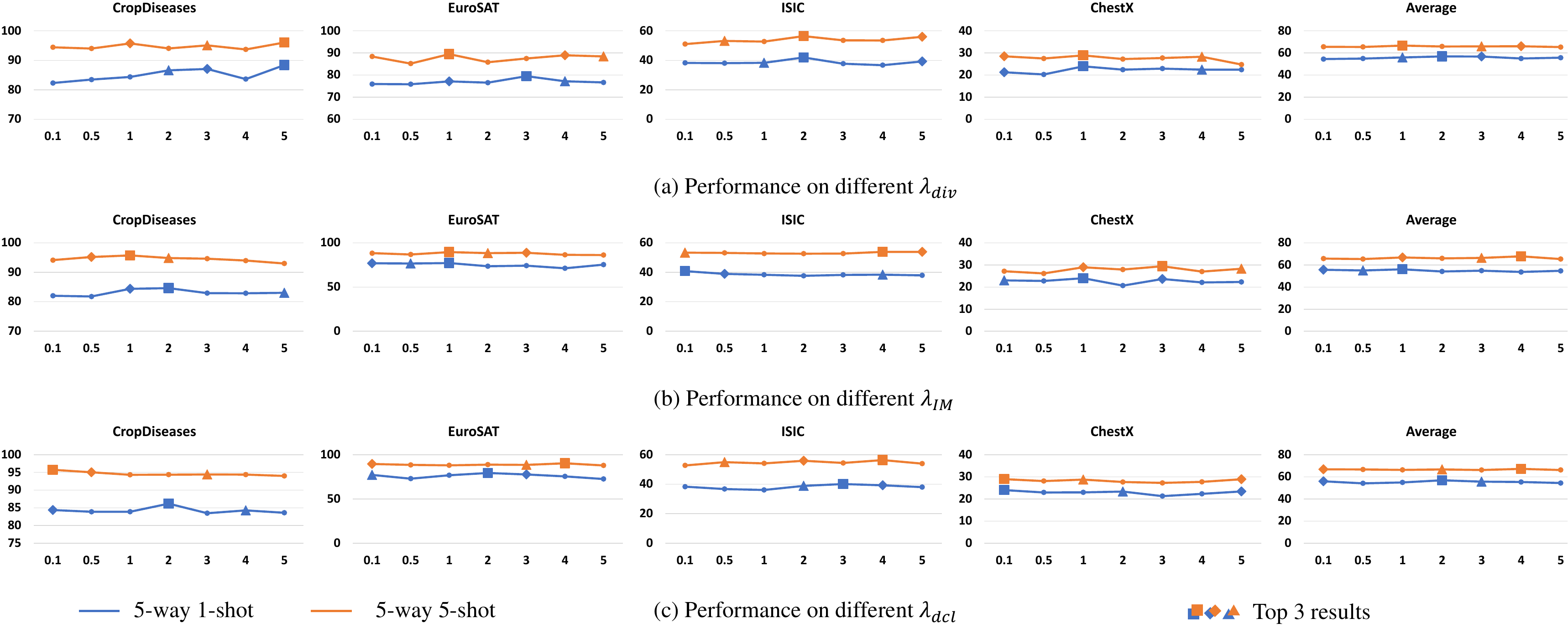}
  \caption{\textcolor{black}{Results of different values of hyperparameters. Square, diamond, and triangle represent the optimal, sub-optimal, and third-optimal results respectively.}}
  \label{parameter_select}
  \vspace{-0.3cm}
\end{figure*}
}

\textcolor{black}{
Examining the presented figure, we draw attention to the top three results achieved for each task across the datasets. Analyzing the averaged outcomes, we observe that the most favorable choices for $\lambda_{div}$ are 1 and 3. However, when $\lambda_{div}$ is set to 3, the model excels in the near-domain but falters in the distant domain when compared to the setting with $\lambda_{div}$ equal to 1. Consequently, a value of 1 is recommended for $\lambda_{div}$. Regarding $\lambda_{IM}$, the optimal selection, as indicated by the average results, is 1. As for $\lambda_{dcl}$, both 0.1 and 2 demonstrate promising performance. Nonetheless, when $\lambda_{dcl}$ is set to 2, it excels in the 5W1S task but underperforms in the 5W5S task. Conversely, setting it to 0.1 yields strong results in both the 5W1S and 5W5S tasks.}

Furthermore, this paper defines $\lambda_{\mathcal{N}}$ (in Eq.~\ref{sup}) as a dynamic hyperparameter that evolves with the training process. Table~\ref{ln} displays the impact of different settings $\lambda_{\mathcal{N}}$ on the results. We evaluate two stable values of $\lambda_{\mathcal{N}}$ in the adaptation phase, which are the minimum and maximum values, respectively. The minimum $\lambda_{\mathcal{N}}$ value is obtained by $(1+10 \times \frac{h_{i}}{H})^{-5.0}$ when $h_{i}$ is $H$, and the maximum one is got from that when $h_{i}$ is 0. When $\lambda_{\mathcal{N}}$ is set to a stable minimum, DCL focus solely on the positive set, ignoring the influence of the negative set. Consequently, the minimum $\lambda_{\mathcal{N}}$ setting yields the poorest average outcomes of 54.68\% and 65.65\% on the 5W1S and the 5W5S tasks, respectively. In the 5W1S task, it is evident that the maximum $\lambda_{\mathcal{N}}$ yields an optimal result of 84.98\% and 38.45\% on CropDiseases and ISIC, outperforming the 84.37\% and 38.13\% achieved by a variable $\lambda_{\mathcal{N}}$. 
When analyzing the results, it can be deduced that a maximum $\lambda_{\mathcal{N}}$ outperforms a variable $\lambda_{\mathcal{N}}$ in the near domain. However, it usually gets poor performance in the distant domain. In contrast, the variable $\lambda_{\mathcal{N}}$ sacrifices some performance in the near domain but significantly improves results in the distant domain. This trade-off enables it to achieve superior average performance, thereby demonstrating remarkable generalization capabilities.
\begin{table}[!t]
  \caption{Performance comparison of different $\lambda_{\mathcal{N}}$ sets. The results show that the variable $\lambda_{\mathcal{N}}$ achieves optimal performance.}
  \label{ln}
  \scriptsize 
  \centering
  \setlength{\tabcolsep}{1.0mm}{
  \begin{tabular}{l|lccccc}
    \toprule
    \textbf{\textit{K}} &  \textbf{$\lambda_{\mathcal{N}}$} & \textbf{CropDiseases} & \textbf{EuroSAT} & \textbf{ISIC} & \textbf{ChestX} & \textbf{Avg}   \\
    \midrule
    \multirow{3}*{1} & Minimum  & 84.27$\pm$0.63 & 73.90$\pm$0.73 & 38.01$\pm$0.58 & 22.53$\pm$0.69 & 54.68  \\
    & Maximum  & \textbf{84.98$\pm$0.53} & 74.88$\pm$0.92 & \textbf{38.45$\pm$0.75} & 22.86$\pm$0.26 & 55.29    \\
    & Variable  & 84.37$\pm$0.99 & \textbf{77.14$\pm$0.71} & 38.13$\pm$0.57 & \textbf{23.98$\pm$0.79} & \textbf{55.91}    \\
    \midrule
    \multirow{3}*{5} & Minimum   & 94.21$\pm$0.45 & 87.89$\pm$0.44 & 52.52$\pm$0.53 & 27.96$\pm$0.73 & 65.65   \\
    & Maximum  & \textbf{95.80$\pm$0.42} & \textbf{89.72$\pm$0.93} & 52.38$\pm$0.35 & 28.20$\pm$0.49 & 66.53    \\
    & Variable  & 95.73$\pm$0.38 & 89.47$\pm$0.42 & \textbf{52.74$\pm$0.69} & \textbf{28.93$\pm$0.41} & \textbf{66.72}    \\
    \bottomrule
  \end{tabular}
  }
  \vspace{-0.3cm}
\end{table}

\subsection{Performance Analysis}
This section includes 3 main stages. Firstly, we examine the role of data augmentation and transduction strength for the IM-DCL performance. Subsequently, we evaluate the effects of varying source models and their quality on IM-DCL. Conclusively, we explore the capacity of IM-DCL to tackle SF-CDFSL without fine-tuning the source model, which is compelling as it indicates that existing pre-trained models can solve CDFSL even with their existing parameters, thereby boosting the learning efficiency of CDFSL and easing its practical implementation requirements.

\subsubsection*{\bf Effect of Data Augmentation}
Existing CDFSL methods typically utilize data augmentation during the fine-tuning phase to alleviate the performance drop due to the limited labeled data. In light of this, we quantify the influence of data augmentation (DA) for IM-DCL. As illustrated in Table~\ref{da}, compared to the results without DA, IM-DCL with DA obtains the superior average results in both the 5W1S (55.91\% vs. 54.29\%) and the 5W5S tasks (66.72\% vs. 65.05\%). Furthermore, the advantage of DA is obvious in the 5W1S and 5W5S outcomes across each dataset, consistent with the conclusions of existing research.
\begin{table}[!t]
  \caption{Effect of data augmentation (DA) on 5-way $K$-shot task.}
  \label{da}
  \scriptsize 
  \centering
  \setlength{\tabcolsep}{1.0mm}{
  \begin{tabular}{l|lccccc}
    \toprule
    \textbf{\textit{K}} &  \textbf{Methods} & \textbf{CropDiseases} & \textbf{EuroSAT} & \textbf{ISIC} & \textbf{ChestX} & \textbf{Avg}   \\
    \midrule
    \multirow{2}*{1} & IM-DCL \textit{w/o DA}  & 82.81$\pm$0.59 & 73.90$\pm$0.51 & 37.81$\pm$0.57 & 22.64$\pm$0.42 & 54.29   \\
    & IM-DCL & \textbf{84.37$\pm$0.99} & \textbf{77.14$\pm$0.71} & \textbf{38.13$\pm$0.57} & \textbf{23.98$\pm$0.79} & \textbf{55.91}    \\
    \midrule
    \multirow{2}*{5} & IM-DCL \textit{w/o DA}  & 93.66$\pm$0.31 & 86.82$\pm$0.72 & 52.03$\pm$0.62 & 27.74$\pm$0.81 & 65.06   \\
    & IM-DCL & \textbf{95.73$\pm$0.38} & \textbf{89.47$\pm$0.42} & \textbf{52.74$\pm$0.69} & \textbf{28.93$\pm$0.41} & \textbf{66.72}    \\
    \bottomrule
  \end{tabular}
  }
  \vspace{-0.3cm}
\end{table}

\subsubsection*{\bf Verification of Transduction Strength}
Currently, several SOTA approaches deploy transductive strategies like label propagation (LP) during the testing phase on the query set~\cite{liu2020feature}. Moreover, existing methods exploit the transductive strategy in the model optimization stage~\cite{li2022ranking,li2023knowledge}. However, heavy reliance on this can cause overfitting to the query set. To assess if the transductive strategy of IM-DCL is overfitting, we use LP in testing and compare its results as an overfitting indicator. Ideally, a transductive model should narrow the gap between its results and LP outcomes while improving performance.

\begin{figure*}[!t]
  \centering
  \includegraphics[width=0.9\textwidth]{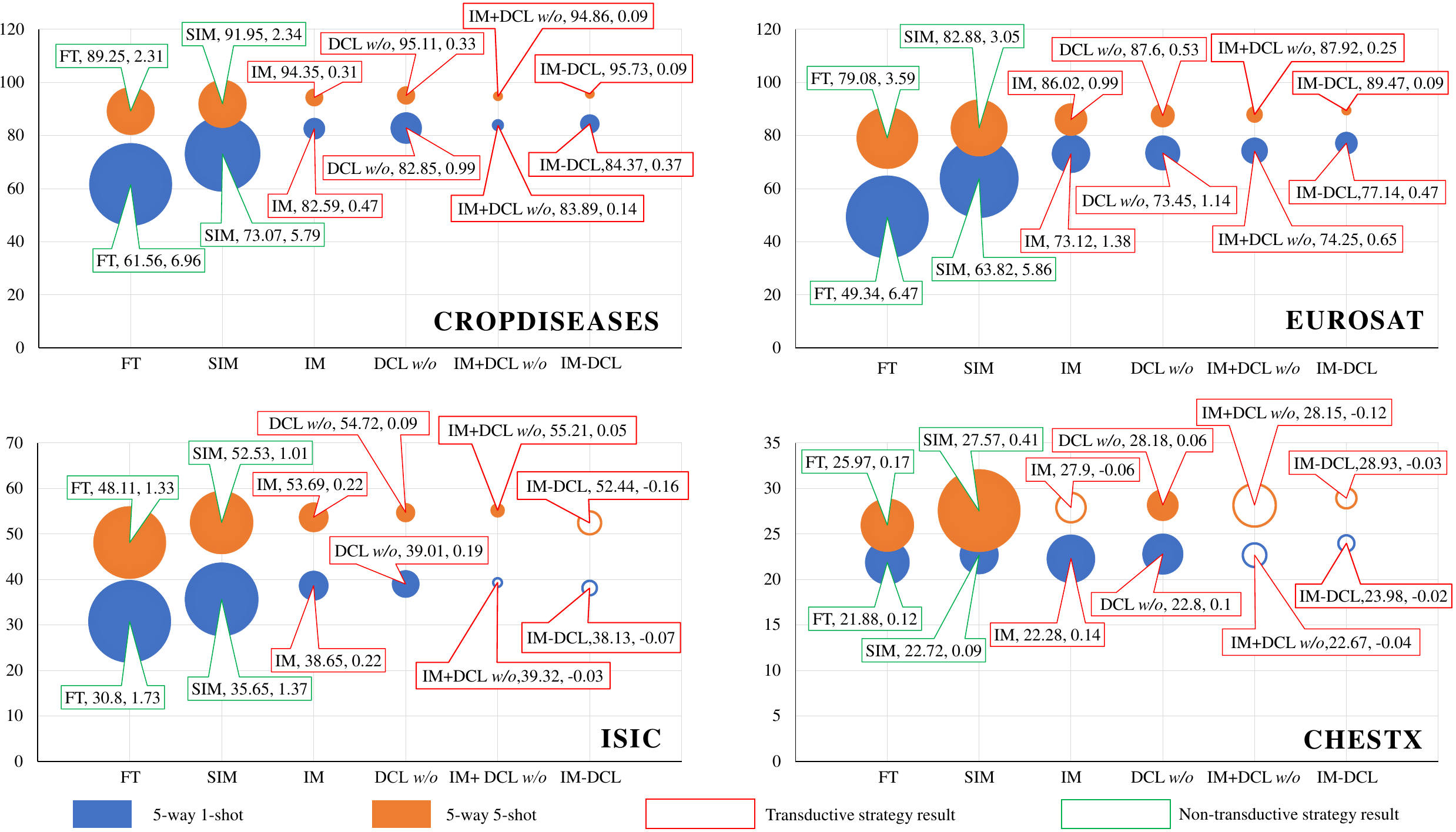}
  \caption{Strength of transduction in IM-DCL. The Figure shows that IM-DCL achieves optimal performance while taking full advantage of the transduction strategy. The size of the circles in the figure indicates the difference between the results enhanced by the LP test and the predicted results. Filled circles indicate that the LP result is higher than the predicted result, while open circles indicate that the LP result is lower than the predicted result. Blue and orange represent 5W1S and 5W5S tasks, respectively. The red/green boxes indicate that the method strategy with/without the transduction mechanism. The contents in the box indicate the strategy, predicted result, and the difference between the predicted and LP result, respectively.}
  \label{lp}
  \vspace{-0.3cm}
\end{figure*}

We verify the efficacy and potential overfitting of transductive measures in IM-DCL. The results are shown in Figure~\ref{lp}, which presents an evaluation of the transductive strength in IM-DCL by exhibiting the disparities between the LP results and the IM-DCL results on 4 datasets.
Firstly, the results on the 4 datasets reveal that leveraging transductive strategies such as `IM', `DCL $w/o$', `IM+DCL $w/o$', and `IM-DCL' typically show the superior performance, in contrast to the `Fine-tuning' and `SIM' that without transduction. Secondly, due to the dual demands of performance and generalization for transduction, the proposed IM-DCL attempts to harmonize high performance with maintaining transduction strength. The figure discloses that compared to `IM+DCL $w/o$', `IM-DCL' obtains the optimal balance between performance and transductive strength, which means that the transductive strategy maximizes its effectiveness while avoiding overfitting on the query set. Moreover, different from the near domain outcomes (CropDiseases and EuroSAT), the results with the transductive strategy surpass the LP results (represented by open circles) on the distant domain task (ISIC and ChestX), highlighting its advantage on the distant domain. In general, `IM-DCL' sufficiently and rationally utilizes the advantages of the transduction strategy and maintains high performance on SF-CDFSL.

\textcolor{black}{
Besides, we evaluate the proposed IM-DCL on the additional query set to verify the generalization ability of IM-DCL. In addition to the required 15 samples in query set, the extra 70 samples are evaluated to illustrate the generalization performance of IM-DCL. Figure~\ref{addition1} indicates the results on the extra 70 samples. As shown in Figure~\ref{addition1}, compared to baseline, the evaluation on the additional query set obtains the excellent 54.8\% and 66.4\% average results for 5W1S and 5W5S, respectively. Compared with the transductive manner, which achieved 55.91\% and 66.72\% on 5W1S and 5W5S respectively, the additional query set samples only reduced the performance by 0.11\% and 0.32\%, reflecting the excellent generalization performance of the proposed IM-DCL. It is worth noting that the result of the additional query set on ISIC (53.56\% for 5W5S) is slightly higher than the transductive result (52.74\%). The analysis believes that when the training data increases, the interference of background factors on ISIC is more likely to be Solved, at this time the transduction strategy is more likely to lead to overfitting.
\begin{figure}
  \centering
  \vspace{-0.2cm}
  \includegraphics[width=0.42\textwidth]{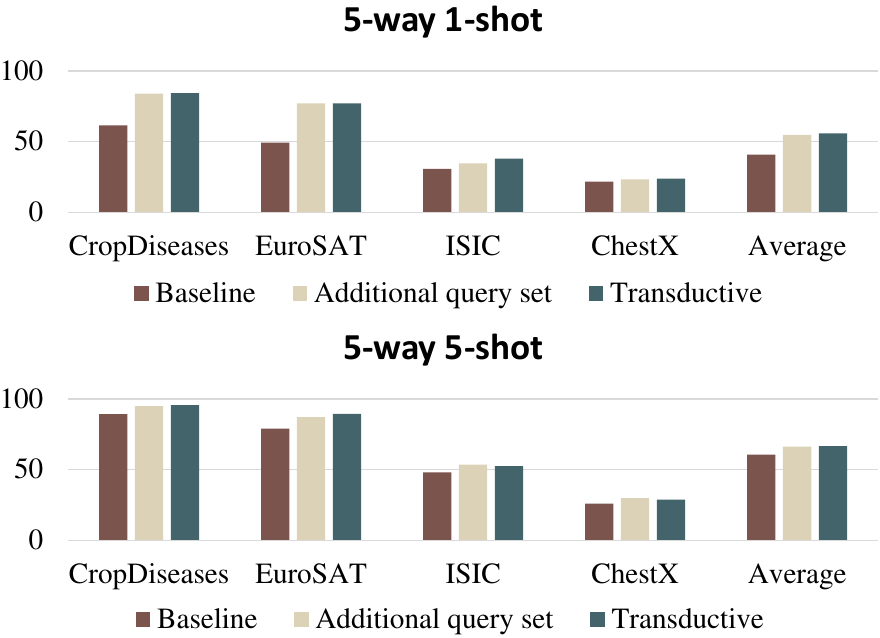}
  \vspace{-0.2cm}
  \caption{\textcolor{black}{Comparison of baseline, performance on the additional query set, and the transductive results.}}
  \label{addition1}
  \vspace{-0.4cm}
\end{figure}
}

\subsubsection*{\bf Effect of Different Source Model}
Besides evaluating the CDFSL performance of IM-DCL with a source model trained via the vanilla CE loss, this study explores the IM-DCL results on other existing visual pre-training models. We first illustrate the robustness of IM-DCL with BSR model~\cite{liu2020feature}, which also utilizes the ResNet10 architecture. Furthermore, the ViT model~\cite{han2022survey} is introduced to evaluate IM-DCL results. 

The baseline and IM-DCL outcomes with different source models are shown in Table~\ref{dsm}, including a standard CE loss-based model (CE), the BSR model, and ViT (`Freeze' is set as `N'). Analysis of the baseline results across the models in the 5W1S task indicates that compared to the CE and BSR models, ViT demonstrates remarkable performance in the near domain (CropDiseases), with an impressive 76.22\%. However, it does not show outstanding performance than BSR in the distant domain (EuroSAT, ISIC and ChestX), securing 54.29\%, 32.19\% and 21.90\% respectively. Analysis believes that this phenomenon is attributed to the deep network architecture of ViT, which makes the model more prone to overfitting in optimization with limited supervision, especially in distant domain tasks. ViT outperforms the other two models in the 5W5S task also validates this analysis. 

We can know that, compared to the baselines, IM-DCL results on all backbones are greatly improved, \ie, IM-DCL weakens the dependence of CDFSL tasks on the source model. Besides, it is worth noting that on CropDiseases, IM-DCL-based BSR achieves the excellent results of 86.76\% and 95.98\% on 5W1S and 5W5S tasks, respectively. While on the other 3 datasets, BSR performance is lower than that of the CE model, which shows that the combination of IM-DCL and the training strategy on the source domain does not ensure a double performance gain.
\begin{table*}
  \caption{Comparison of the performance with different source models on the 5W1S and 5W5S tasks. IM-DCL improves the results on almost all datasets with each backbone. `Model' means the source model. `Freeze' means if the source model is frozen, in which `N' means the source model can be fine-tuned, and `Y' represents the source model is frozen.}
  \label{dsm}
  \scriptsize 
  \centering
  \setlength{\tabcolsep}{1.0mm}{
  \begin{tabular}{clc|ccccc|ccccc}
    \toprule
     \multirow{2}*{\textbf{Freeze}} & \multirow{2}*{\textbf{Methods}} & \multirow{2}*{\textbf{Model}}  & \multicolumn{5}{c|}{\textbf{5W1S}}  & \multicolumn{5}{c}{\textbf{5W5S}}   \\
    & &  & \textbf{CropDiseases} & \textbf{EuroSAT} & \textbf{ISIC} & \textbf{ChestX} & \textbf{Avg} & \textbf{CropDiseases} & \textbf{EuroSAT} & \textbf{ISIC} & \textbf{ChestX} & \textbf{Avg}   \\
    \midrule
    \multirow{6}*{N} & \multirow{3}*{Baseline} & CE & 61.56$\pm$0.90 & 49.34$\pm$0.85 & 30.80$\pm$0.59 &  21.88$\pm$0.38 & 40.90 & 89.25$\pm$0.51 & 79.08$\pm$0.61 & 48.11$\pm$0.64 & 25.97$\pm$0.41 & 60.60    \\
    & & BSR & 59.39$\pm$0.83 & 56.44$\pm$0.76 & 32.38$\pm$0.61 & 22.87$\pm$0.39 & 42.77 & 89.07$\pm$0.51 & 81.86$\pm$0.56 & 51.78$\pm$0.66 & 27.68$\pm$0.45 & 62.60   \\
    & & ViT & 76.22$\pm$0.89 & 54.29$\pm$0.89 & 32.19$\pm$0.57 & 21.90$\pm$0.39 & 46.15 & 95.12$\pm$0.40 & 81.65$\pm$0.71 & 53.00$\pm$0.66 & 25.58$\pm$0.45 & 63.84   \\
    \cline{2-13}
    & \multirow{3}*{ IM-DCL} & CE & 84.37$\pm$0.99 & 77.14$\pm$0.71 & 38.13$\pm$0.57 & 23.98$\pm$0.79 & 55.91 & 95.73$\pm$0.38 & 89.47$\pm$0.42 & 52.74$\pm$0.69 & 28.93$\pm$0.41  & 66.72   \\
    & & BSR & 86.76$\pm$0.52 & 69.92$\pm$0.95 & 37.05$\pm$0.64 & 22.35$\pm$0.51 & 54.02 & 95.98$\pm$0.39 & 87.16$\pm$0.83 & 55.65$\pm$0.63 & 28.26$\pm$0.64 & 66.76   \\
    & & ViT & 85.98$\pm$0.72 & 74.68$\pm$0.86 & 38.20$\pm$0.75 & 22.68$\pm$0.27 & 55.39 & 95.93$\pm$0.50 & 87.76$\pm$0.57 & 53.89$\pm$0.35 & 27.98$\pm$0.46 & 66.39   \\
    \midrule
    \midrule
    \multirow{12}*{Y} & \multirow{3}*{Baseline} & CE & 59.77$\pm$0.86 & 48.72$\pm$0.93 & 30.66$\pm$0.64 & 21.30$\pm$0.43 & 40.11 & 87.48$\pm$0.58 & 75.69$\pm$0.66 & 43.56$\pm$0.60 & 25.35$\pm$0.96 & 58.02    \\
    & & BSR & 59.74$\pm$0.88 & 54.13$\pm$0.81 & 32.62$\pm$0.58 & 21.90$\pm$0.39 & 42.10 & 81.61$\pm$0.66 & 74.54$\pm$0.70 & 43.24$\pm$0.61 & 24.71$\pm$0.43 & 56.03  \\
    & & ViT & 67.07$\pm$1.01 & 57.26$\pm$0.86 & 27.34$\pm$0.48 & 20.28$\pm$0.28 & 42.99 & 89.49$\pm$0.55 & 77.30$\pm$0.53 & 40.87$\pm$0.59 & 22.34$\pm$0.39 & 57.50   \\
    \cline{2-13}
    & \multirow{3}*{Prompt Baseline} & CE & 64.98$\pm$0.86 & 58.00$\pm$0.87 & 32.68$\pm$0.61 & 22.50$\pm$0.42 & 44.54 & 90.68$\pm$0.49 & 81.65$\pm$0.61 & 46.91$\pm$0.62 & 26.18$\pm$0.43 & 61.51   \\
    & & BSR & 61.52$\pm$0.87 & 55.24$\pm$0.78 & 31.79$\pm$0.60 & 22.08$\pm$0.39 & 42.66 & 83.33$\pm$0.64 & 74.66$\pm$0.70 & 42.71$\pm$0.60 & 24.67$\pm$0.43 & 56.34   \\
    & & ViT & 63.65$\pm$0.85 & 56.25$\pm$0.86 & 35.80$\pm$0.55 & 21.33$\pm$0.37 & 44.27 & 88.28$\pm$0.96 & 45.76$\pm$0.91 & 37.43$\pm$0.55 & 21.98$\pm$0.39 & 48.36   \\
    \cline{2-13}
    & \multirow{3}*{ IM-DCL} & CE & 74.23$\pm$0.91 & 69.91$\pm$0.92 & 34.23$\pm$0.65 & 22.89$\pm$0.43 & 50.32 & 88.68$\pm$0.58 & 82.90$\pm$0.85 & 45.58$\pm$0.83 & 25.98$\pm$0.47 & 60.79   \\
    & & BSR & 61.77$\pm$0.95 & 54.68$\pm$0.80 & 32.36$\pm$0.58 & 22.29$\pm$0.40 & 42.78 & 80.98$\pm$0.80 & 74.66$\pm$0.69 & 42.65$\pm$0.62 & 24.87$\pm$0.52 & 55.79   \\
    & & ViT & 69.49$\pm$0.94 & 57.80$\pm$0.57 & 27.29$\pm$0.68 & 20.43$\pm$0.20 & 43.75 & 88.85$\pm$0.50 & 76.35$\pm$0.68 & 40.19$\pm$0.57 & 22.75$\pm$0.57 & 57.04   \\
    \cline{2-13}
    & \multirow{3}*{Prompt IM-DCL} & CE & 74.04$\pm$0.89 & 71.84$\pm$0.97 & 35.24$\pm$0.72 & 23.08$\pm$0.70 & 51.05 & 88.69$\pm$0.58 & 83.76$\pm$0.67 & 47.12$\pm$0.46 & 26.79$\pm$0.82 & 61.63   \\
    & & BSR & 62.49$\pm$0.88 & 56.10$\pm$0.62 & 31.44$\pm$0.58 & 21.75$\pm$0.30 & 42.95 & 82.73$\pm$0.55 & 75.13$\pm$0.66 & 42.66$\pm$0.60 & 25.00$\pm$0.47 & 56.38   \\
    & & ViT & 69.78$\pm$0.78 & 56.94$\pm$0.70 & 26.63$\pm$0.42 & 21.20$\pm$0.32 & 43.64 & 88.51$\pm$0.91 & 72.48$\pm$0.64 & 36.27$\pm$0.79 & 21.98$\pm$0.63 & 54.81   \\
    \bottomrule
  \end{tabular}
  }
  \vspace{-0.3cm}
\end{table*}

\vspace{-10pt}
\textcolor{black}{
\subsubsection*{\bf Effect of Different Source Data}
In addition to the source models pretrained on the ImageNet series datasets, we also utilized two domain-specific datasets, PatternNet~\cite{pattern} and the Skin dataset~\footnote{\url{https://www.cvmart.net/dataSets/detail/450}}, to train the source model. PatternNet~\cite{pattern} is a large-scale high-resolution remote sensing dataset collected for remote sensing image retrieval. There are 38 classes and each class has 800 images of size 256×256 pixels. Skin dataset contains image data of 23 types of skin diseases, and the total number of images is approximately There are 19,500 images, of which about 15,500 have been segmented in the training set and the rest in the test set. Table~\ref{pretrained11} illustrates the comparison of models pretrained on different source data.
\begin{table}[h]
  \caption{\textcolor{black}{Performance of models pretrained on different source datasets `PatternNet' and `Skin'. `K' means 5-way \textit{K}-shot task.}}
  \label{pretrained11}
  \scriptsize 
  \centering
  \setlength{\tabcolsep}{1.0mm}{
  \begin{tabular}{l|l|l|cccc}
    \toprule
     \textit{K} & \textbf{Method} & \textbf{Source data} & \textbf{CropDiseases} & \textbf{EuroSAT} & \textbf{ISIC} & \textbf{ChestX}   \\
    \toprule
    \multirow{6}*{1} & \multirow{3}*{Baseline} & \textit{mini}ImageNet & \textbf{61.56$\pm$0.90} & 49.34$\pm$0.85 & 30.80$\pm$0.59 &  \textbf{21.88$\pm$0.38}  \\
    &  & PatternNet & 61.13$\pm$0.57 & \textbf{68.71$\pm$0.72} & 30.22$\pm$0.11 & 21.02$\pm$0.47   \\
    &  & Skin & 36.81$\pm$0.55 & 34.73$\pm$0.72 & \textbf{37.56$\pm$0.90} & 21.02$\pm$0.60   \\
    \cline{2-7}
     & \multirow{3}*{IM-DCL} & \textit{mini}ImageNet & \textbf{84.37$\pm$0.99} & 77.14$\pm$0.71 & 38.13$\pm$0.57 &  \textbf{23.98$\pm$0.79}   \\
     & & PatternNet & 78.39$\pm$0.78 & \textbf{82.04$\pm$0.95} & 31.82$\pm$0.34 & 21.87$\pm$0.33   \\
     & & Skin & 56.59$\pm$0.33 & 57.24$\pm$0.39 & \textbf{42.71$\pm$0.97} & 21.38$\pm$0.78  \\
     \midrule
     \multirow{6}*{5} & \multirow{3}*{Baseline} & \textit{mini}ImageNet & \textbf{89.25$\pm$0.51} & 79.08$\pm$0.61 & 48.11$\pm$0.64 & \textbf{25.97$\pm$0.41}  \\
    &  & PatternNet & 87.39$\pm$0.88 & \textbf{86.76$\pm$0.75} & 49.29$\pm$0.64 & 25.04$\pm$0.53   \\
    &  & Skin & 56.79$\pm$0.51 & 57.53$\pm$0.96 & \textbf{55.53$\pm$0.36} & 25.24$\pm$0.23  \\
    \cline{2-7}
     & \multirow{3}*{IM-DCL} & \textit{mini}ImageNet & \textbf{95.73$\pm$0.38} & 89.47$\pm$0.42 & 52.74$\pm$0.69 & \textbf{28.93$\pm$0.41}  \\
     & & PatternNet & 90.22$\pm$0.47 & \textbf{91.80$\pm$0.42} & 51.47$\pm$0.86 & 26.40$\pm$0.35  \\
     & & Skin & 69.87$\pm$0.88 & 64.22$\pm$0.74 & \textbf{59.82$\pm$0.28} & 27.44$\pm$0.93  \\
    \bottomrule
  \end{tabular}
  }
   \vspace{-0.4cm}
\end{table}
}

\textcolor{black}{Compared to \textit{mini}ImageNet source model, PatternNet an Skin source models obtain the optimal performance (68.71\% and 37.56\% for 5W1S task , while 86.76\% and 55.53\% for 5W5S task) on EuroSAT and ISIC, respectively. However, miniImageNet still get the optimal results on the other datasets. Analysis believes that \textit{mini}ImageNet is larger than the latter two, resulting in stronger generalization of its corresponding source model. In addition, compared to baseline, IM-DCL improves the performance on all sources.}

\subsubsection*{\bf Effect of Source Model Quality}
\begin{figure}[!t]
  \centering
  \includegraphics[width=0.45\textwidth]{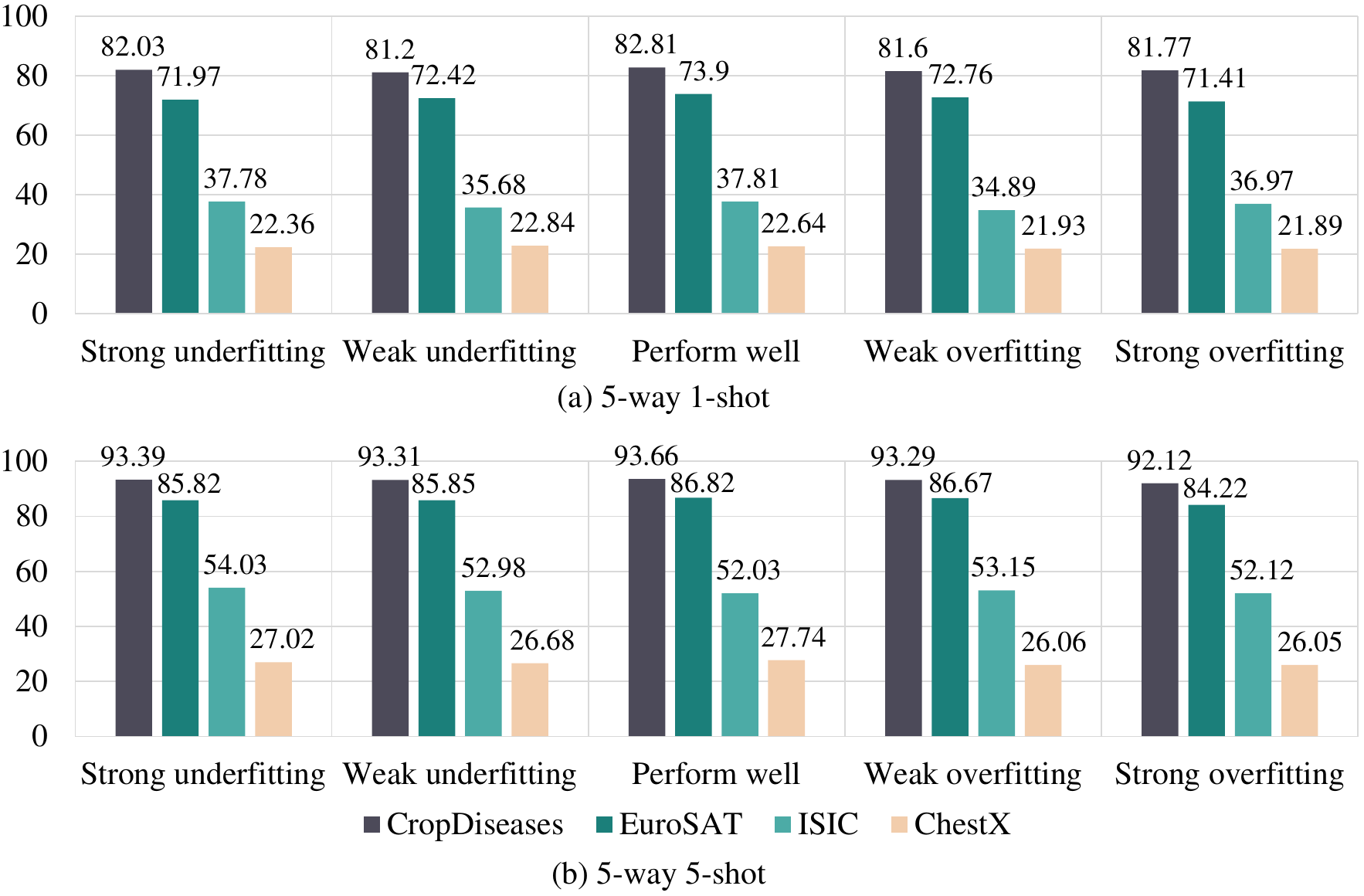}
  \caption{Comparison of the IM-DCL performance about the different phase of source models on (a) the 5W1S and (b) the 5W5S tasks. The results of IM-DCL on the five different models tend to be balanced, which shows that on the different quality models, IM-DCL can still maximize the use of source model knowledge.}
  \label{phase}
  \vspace{-0.5cm}
\end{figure}

We test the performance of source models of varying quality, given their origin from pre-trained models. We use 5 CE models, simulating different qualities: strong underfitting, weak underfitting, optimal, weak overfitting, and strong overfitting. Beyond the optimal epoch 1200, we use parameters from strong underfitting (epoch 400), weak underfitting (epoch 800), weak overfitting (epoch 1600), and strong overfitting (epoch 2000). Figure~\ref{phase} shows the IM-DCL results for these 5 models.

The results of Figure~\ref{phase} reveal that, in the 5W1S task, compared to the optimal one, the performance differences of the underfitting models not exceeding 1.61\%, 1.48\%, 2.13\%, and 0.2\% on the 4 datasets, respectively. Similarly, with an overfitting model, the performance bias remains confined within 1.21\%, 2.49\%, 2.92\%, and 0.75\% on 4 datasets. IM-DCL produces performance differentials of no more than 1.54\%, 2.60\%, 2.00\%, and 1.69\% under varying overfitting or underfitting models in the 5W5S task. These results highlight the robustness of IM-DCL, maintaining reliable outcomes even under the underfitting and overfitting source models. Moreover, these performance variations highlight the impact of model quality for the distant domain task, \ie, the performance drop in the distant domain of underfitting or overfitting models is notably more pronounced than in near domains.

\subsubsection*{\bf Effect of Frozen Source Model}
\label{freezes}
Beyond simply fine-tuning the source model, a key direction in SF-CDFSL is addressing the CDFSL challenge with the model's fixed parameters. This approach is driven by the high computational costs of fine-tuning large foundational models. If a source model can tackle CDFSL without changing parameters, it eliminates the need for a smaller architecture to prevent overfitting due to limited labeled data in the target domain. In this section, we present results with the source model frozen. Inspired by prompt learning~\cite{jia2022visual}, we use a visual prompt to enhance CDFSL performance. We create a prompt matching the sample size and concatenate it with each sample width-wise, forming new inputs. Validation results are in Table~\ref{dsm} (`Freeze' is set as `Y').

We first exhibit and analyze the baseline and IM-DCL results in Table~\ref{dsm}. Firstly, the results between the freeze and the fine-tuned model exhibit that both baseline and IM-DCL achieve unsatisfactory results on all frozen backbones. Although the introduction of the visual prompt brings some promising progress, the performance is still lower than that of fine-tuned backbones. Therefore, it is still a challenging problem to directly utilize the frozen source model parameters to solve the CDFSL problem. Next, we analyze the results of frozen backbones. First of all, in the 5W1S task, IM-DCL surpasses the baseline in average performance across all models.
In the 5W5S task, IM-DCL performs the average result of 60.79\% that better which of 58.02\% for the Baseline in the CE model, while obtaining 55.79\% and 57.04\% average lower scores in BSR and ViT. The analysis argues that IM-DCL is designed for the lack of labeled samples, making up for the shallow network processing limitations in the 5W1S task. As the volume of data increases, the deep network architecture (such as ViT) exhibits strong processing capabilities. While the batch spectral regularization mechanism in the training phase helps the BSR model to handle the target domain tasks. 

Secondly, comparing the results of the Baseline and Prompt baseline, we can know that a basic visual prompt can enhance performance on shallow networks.
For instance, in both the 5W1S and 5W5S tasks, the average results of the CE and BSR models are increased. 
However, in ViT, the introduction of visual prompts reduces performance in near domain tasks. In contrast, in distant domain, the ViT introduced by prompt has significant improvement, like an increase from 27.34\% to 35.80\% on ISIC and 20.28\% to 21.33\% on ChestX
. This represents that visual prompts greatly help distant domain tasks in deep networks, although this advantage diminishes as data volumes grow (in the 5W5S task, the introduction of visual prompts does not get positive effects). From our analysis, the design and integration manner of the visual prompt is crucial to the final performance with varying impacts on different models, which have the same conclusion with~\cite{jia2022visual}. Lastly, the result of Prompt IM-DCL shows that CE and BSR achieve better average results on the 5W1S and 5W5S tasks than IM-DCL. However, the improvement of prompt on ViT is not pronounced. In the 5W1S and 5W5S tasks, Prompt IM-DCL registered lower average results of 43.64\% and 54.81\%, respectively. This further illustrates the design and exploitation manner of basic visual prompts are not matched ViT, compromising its performance.

\vspace{-0.6cm}
\textcolor{black}{
\section{Limitations and Future Work}
While the proposed method demonstrates promising potential in SF-CDFSL, there are several limitations that warrant further consideration. Firstly, one limitation of our proposed method is its sub-optimal performance on datasets that are susceptible to background factors, such as ISIC. The method currently lacks mechanisms to effectively counteract the influence of background noise, which can lead to decreased accuracy. Future enhancements will focus on integrating strategies to mitigate background interference. This could include the incorporation of interpretability or attention mechanisms that enable the model to focus more on foreground classification while disregarding background information.
}

\textcolor{black}{In addition, while our method demonstrates some improvement on ChestX compared to other approaches, it still falls short of effectively addressing the challenges inherent to this task. The inherent complexity of X-ray image analysis and the large domain gap with natural images constitute significant obstacles that current methods cannot fully overcome. In future work, we aim to explore more effective methods within the SF-CDFSL framework to enhance performance in X-ray FSL image recognition tasks. Potential avenues for improvement include the integration of multi-modal information, the usage of large vision models. It could provide richer context and help the model to better differentiate and classify nuanced features in medical imaging. By addressing these challenges, we hope to significantly advance the effectiveness of FSL in complex medical imaging scenarios.
}

\section{Conclusion}
This paper addresses the Source-Free Cross-Domain Few-Shot Learning (SF-CDFSL) challenge, where CDFSL is tackled without accessing source data. We introduce the Enhanced Information Maximization with Distance-Aware Contrastive Learning (IM-DCL) to address FSL in the target domain using minimal supervised data. IM-DCL employs an enhanced information maximization (IM) strategy and a transductive mechanism. IM ensures the source model produces predictions with individual certainty and global diversity. To counter the limitation of IM in defining the correct decision boundary, we propose a distance-aware contrastive learning (DCL) strategy. In DCL, features are associated with positive and negative sets from a memory bank based on distance, aiming to bring features closer to positive sets and farther from negative ones. 
Evaluations on four BSCD-FSL datasets show IM-DCL outperforms adapted strategy-driven CDFSL methods and is competitive with training strategy-based approaches. The efficacy of IM-DCL on frozen source models also suggests promising avenues for future research.

\vspace{-0.1cm}

\normalem
\bibliographystyle{IEEEtran}

\bibliography{IEEEabrv,ref}

\begin{thebibliography}{10}
\providecommand{\url}[1]{#1}
\csname url@samestyle\endcsname
\providecommand{\newblock}{\relax}
\providecommand{\bibinfo}[2]{#2}
\providecommand{\BIBentrySTDinterwordspacing}{\spaceskip=0pt\relax}
\providecommand{\BIBentryALTinterwordstretchfactor}{4}
\providecommand{\BIBentryALTinterwordspacing}{\spaceskip=\fontdimen2\font plus
\BIBentryALTinterwordstretchfactor\fontdimen3\font minus \fontdimen4\font\relax}
\providecommand{\BIBforeignlanguage}[2]{{%
\expandafter\ifx\csname l@#1\endcsname\relax
\typeout{** WARNING: IEEEtran.bst: No hyphenation pattern has been}%
\typeout{** loaded for the language `#1'. Using the pattern for}%
\typeout{** the default language instead.}%
\else
\language=\csname l@#1\endcsname
\fi
#2}}
\providecommand{\BIBdecl}{\relax}
\BIBdecl

\bibitem{lake2015human}
B.~M. Lake, R.~Salakhutdinov, and J.~B. Tenenbaum, ``Human-level concept learning through probabilistic program induction,'' \emph{Science}, vol. 350, no. 6266, pp. 1332--1338, 2015.

\bibitem{wang2020generalizing}
Y.~Wang, Q.~Yao, J.~T. Kwok, and L.~M. Ni, ``Generalizing from a few examples: A survey on few-shot learning,'' \emph{ACM CSUR}, vol.~53, no.~3, pp. 1--34, 2020.

\bibitem{sun2023fastal}
S.~Sun, H.~Xu, Y.~Li, P.~Li, B.~Sheng, and X.~Lin, ``Fastal: Fast evaluation module for efficient dynamic deep active learning using broad learning system,'' \emph{TCSVT}, 2023.

\bibitem{song2023comprehensive}
Y.~Song, T.~Wang, P.~Cai, S.~K. Mondal, and J.~P. Sahoo, ``A comprehensive survey of few-shot learning: Evolution, applications, challenges, and opportunities,'' \emph{ACM CSUR}, 2023.

\bibitem{yang2022efficient}
Z.~Yang, C.~Zhang, R.~Li, Y.~Xu, and G.~Lin, ``Efficient few-shot object detection via knowledge inheritance,'' \emph{TIP}, vol.~32, pp. 321--334, 2022.

\bibitem{vu2023instance}
A.~Vu, T.~Do, N.~Nguyen, V.~Nguyen, T.~Ngo, and T.~Nguyen, ``Instance-level few-shot learning with class hierarchy mining,'' \emph{TIP}, 2023.

\bibitem{guo2022learning}
Y.~Guo, R.~Du, X.~Li, J.~Xie, Z.~Ma, and Y.~Dong, ``Learning calibrated class centers for few-shot classification by pair-wise similarity,'' \emph{TIP}, vol.~31, pp. 4543--4555, 2022.

\bibitem{xu2023deep}
H.~Xu, S.~Zhi, S.~Sun, V.~M. Patel, and L.~Liu, ``Deep learning for cross-domain few-shot visual recognition: A survey,'' \emph{arXiv preprint arXiv:2303.08557}, 2023.

\bibitem{tseng2020cross}
H.~Tseng, H.~Lee, J.~Huang, and M.~Yang, ``Cross-domain few-shot classification via learned feature-wise transformation,'' in \emph{ICLR}, 2020.

\bibitem{phoo2020self}
C.~P. Phoo and B.~Hariharan, ``Self-training for few-shot transfer across extreme task differences,'' in \emph{ICLR}, 2020.

\bibitem{islam2021dynamic}
A.~Islam, C.~R. Chen, R.~Panda, L.~Karlinsky, R.~Feris, and R.~J. Radke, ``Dynamic distillation network for cross-domain few-shot recognition with unlabeled data,'' \emph{NIPS}, vol.~34, pp. 3584--3595, 2021.

\bibitem{xu2022cross}
H.~Xu, S.~Zhi, and L.~Liu, ``Cross-domain few-shot classification via inter-source stylization,'' \emph{ICIP}, 2023.

\bibitem{fu2023styleadv}
Y.~Fu, Y.~Xie, Y.~Fu, and Y.~Jiang, ``Styleadv: Meta style adversarial training for cross-domain few-shot learning,'' in \emph{CVPR}, 2023, pp. 24\,575--24\,584.

\bibitem{zhao2023fs}
Y.~Zhao and N.~Cheung, ``Fs-ban: Born-again networks for domain generalization few-shot classification,'' \emph{TIP}, 2023.

\bibitem{li2022ranking}
P.~Li, S.~Gong, C.~Wang, and Y.~Fu, ``Ranking distance calibration for cross-domain few-shot learning,'' in \emph{CVPR}, 2022, pp. 9099--9108.

\bibitem{zhao2023dual}
Y.~Zhao, T.~Zhang, J.~Li, and Y.~Tian, ``Dual adaptive representation alignment for cross-domain few-shot learning,'' \emph{TPAMI}, 2023.

\bibitem{liang2020we}
J.~Liang, D.~Hu, and J.~Feng, ``Do we really need to access the source data? source hypothesis transfer for unsupervised domain adaptation,'' in \emph{ICML}.\hskip 1em plus 0.5em minus 0.4em\relax PMLR, 2020, pp. 6028--6039.

\bibitem{ding2022source}
N.~Ding, Y.~Xu, Y.~Tang, C.~Xu, Y.~Wang, and D.~Tao, ``Source-free domain adaptation via distribution estimation,'' in \emph{CVPR}, 2022, pp. 7212--7222.

\bibitem{radford2021learning}
A.~Radford, J.~W. Kim, C.~Hallacy, A.~Ramesh, G.~Goh, S.~Agarwal, G.~Sastry, A.~Askell, P.~Mishkin, and J.~Clark, ``Learning transferable visual models from natural language supervision,'' in \emph{ICML}.\hskip 1em plus 0.5em minus 0.4em\relax PMLR, 2021, pp. 8748--8763.

\bibitem{dosovitskiy2020image}
A.~Dosovitskiy, L.~Beyer, A.~Kolesnikov, D.~Weissenborn, X.~Zhai, T.~Unterthiner, M.~Dehghani, M.~Minderer, G.~Heigold, and S.~Gelly, ``An image is worth 16x16 words: Transformers for image recognition at scale,'' in \emph{ICLR}, 2020.

\bibitem{liu2020feature}
B.~Liu, Z.~Zhao, Z.~Li, J.~Jiang, Y.~Guo, and J.~Ye, ``Feature transformation ensemble model with batch spectral regularization for cross-domain few-shot classification,'' \emph{CVPR Challenge}, 2020.

\bibitem{vapnik1999nature}
V.~Vapnik, \emph{The nature of statistical learning theory}.\hskip 1em plus 0.5em minus 0.4em\relax Springer science \& business media, 1999.

\bibitem{gammerman2013learning}
A.~Gammerman, V.~Vovk, and V.~Vapnik, ``Learning by transduction,'' \emph{arXiv preprint arXiv:1301.7375}, 2013.

\bibitem{guo2020broader}
Y.~Guo, N.~C. Codella, L.~Karlinsky, J.~V. Codella, J.~R. Smith, K.~Saenko, T.~Rosing, and R.~Feris, ``A broader study of cross-domain few-shot learning,'' in \emph{ECCV}.\hskip 1em plus 0.5em minus 0.4em\relax Springer, 2020, pp. 124--141.

\bibitem{sun2021explanation}
J.~Sun, S.~Lapuschkin, W.~Samek, Y.~Zhao, N.~Cheung, and A.~Binder, ``Explanation-guided training for cross-domain few-shot classification,'' in \emph{ICPR}.\hskip 1em plus 0.5em minus 0.4em\relax IEEE, 2021, pp. 7609--7616.

\bibitem{fu2021meta}
Y.~Fu, Y.~Fu, and Y.~Jiang, ``Meta-fdmixup: Cross-domain few-shot learning guided by labeled target data,'' in \emph{ACM MM}, 2021, pp. 5326--5334.

\bibitem{fu2022generalized}
Y.~Fu, Y.~Fu, J.~Chen, and Y.~Jiang, ``Generalized meta-fdmixup: Cross-domain few-shot learning guided by labeled target data,'' \emph{TIP}, vol.~31, pp. 7078--7090, 2022.

\bibitem{fu2022wave}
Y.~Fu, Y.~Xie, Y.~Fu, J.~Chen, and Y.~Jiang, ``Wave-san: Wavelet based style augmentation network for cross-domain few-shot learning,'' \emph{arXiv preprint arXiv:2203.07656}, 2022.

\bibitem{zhang2022free}
J.~Zhang, J.~Song, L.~Gao, and H.~Shen, ``Free-lunch for cross-domain few-shot learning: Style-aware episodic training with robust contrastive learning,'' in \emph{ACM MM}, 2022, pp. 2586--2594.

\bibitem{graph1}
Y.~Zhang, W.~Li, M.~Zhang, S.~Wang, R.~Tao, and Q.~Du, ``Graph information aggregation cross-domain few-shot learning for hyperspectral image classification,'' \emph{TNNLS}, pp. 1--14, 2022.

\bibitem{single1}
Y.~Zhang, W.~Li, W.~Sun, R.~Tao, and Q.~Du, ``Single-source domain expansion network for cross-scene hyperspectral image classification,'' \emph{TIP}, vol.~32, pp. 1498--1512, 2023.

\bibitem{topological1}
Y.~Zhang, W.~Li, M.~Zhang, Y.~Qu, R.~Tao, and H.~Qi, ``Topological structure and semantic information transfer network for cross-scene hyperspectral image classification,'' \emph{TNNLS}, 2021.

\bibitem{pei2023uncertainty}
J.~Pei, Z.~Jiang, A.~Men, L.~Chen, Y.~Liu, and Q.~Chen, ``Uncertainty-induced transferability representation for source-free unsupervised domain adaptation,'' \emph{TIP}, vol.~32, pp. 2033--2048, 2023.

\bibitem{roy2022uncertainty}
S.~Roy, M.~Trapp, A.~Pilzer, J.~Kannala, N.~Sebe, E.~Ricci, and A.~Solin, ``Uncertainty-guided source-free domain adaptation,'' in \emph{ECCV}.\hskip 1em plus 0.5em minus 0.4em\relax Springer, 2022, pp. 537--555.

\bibitem{xia2021adaptive}
H.~Xia, H.~Zhao, and Z.~Ding, ``Adaptive adversarial network for source-free domain adaptation,'' in \emph{ICCV}, 2021, pp. 9010--9019.

\bibitem{yang2021exploiting}
S.~Yang, J.~van~de Weijer, L.~Herranz, and S.~Jui, ``Exploiting the intrinsic neighborhood structure for source-free domain adaptation,'' \emph{NIPS}, vol.~34, pp. 29\,393--29\,405, 2021.

\bibitem{liu2018learning}
Y.~Liu, J.~Lee, M.~Park, S.~Kim, E.~Yang, S.~Hwang, and Y.~Yang, ``Learning to propagate labels: Transductive propagation network for few-shot learning,'' in \emph{ICLR}, 2019.

\bibitem{qiao2019transductive}
L.~Qiao, Y.~Shi, J.~Li, Y.~Wang, T.~Huang, and Y.~Tian, ``Transductive episodic-wise adaptive metric for few-shot learning,'' in \emph{ICCV}, 2019, pp. 3603--3612.

\bibitem{ma2020transductive}
Y.~Ma, S.~Bai, S.~An, W.~Liu, A.~Liu, X.~Zhen, and X.~Liu, ``Transductive relation-propagation network for few-shot learning.'' in \emph{IJCAI}, vol.~20, 2020, pp. 804--810.

\bibitem{boudiaf2020information}
M.~Boudiaf, I.~Ziko, J.~Rony, J.~Dolz, P.~Piantanida, and I.~Ben~Ayed, ``Information maximization for few-shot learning,'' \emph{NIPS}, vol.~33, pp. 2445--2457, 2020.

\bibitem{krause2010discriminative}
A.~Krause, P.~Perona, and R.~Gomes, ``Discriminative clustering by regularized information maximization,'' \emph{NIPS}, vol.~23, 2010.

\bibitem{hu2017learning}
W.~Hu, T.~Miyato, S.~Tokui, E.~Matsumoto, and M.~Sugiyama, ``Learning discrete representations via information maximizing self-augmented training,'' in \emph{ICML}.\hskip 1em plus 0.5em minus 0.4em\relax PMLR, 2017, pp. 1558--1567.

\bibitem{yang2022attracting}
S.~Yang, Y.~Wang, K.~Wang, and S.~Jui, ``Attracting and dispersing: A simple approach for source-free domain adaptation,'' in \emph{NIPS}, 2022.

\bibitem{thoma2020soft}
J.~Thoma, D.~P. Paudel, and L.~V. Gool, ``Soft contrastive learning for visual localization,'' \emph{NIPS}, vol.~33, pp. 11\,119--11\,130, 2020.

\bibitem{liang2021domain}
J.~Liang, D.~Hu, and J.~Feng, ``Domain adaptation with auxiliary target domain-oriented classifier,'' in \emph{CVPR}, 2021, pp. 16\,632--16\,642.

\bibitem{yang2021generalized}
S.~Yang, Y.~Wang, J.~Van De~Weijer, L.~Herranz, and S.~Jui, ``Generalized source-free domain adaptation,'' in \emph{ICCV}, 2021, pp. 8978--8987.

\bibitem{mohanty2016using}
S.~P. Mohanty, D.~P. Hughes, and M.~Salath{\'e}, ``Using deep learning for image-based plant disease detection,'' \emph{Front. Plant Sci.}, vol.~7, p. 1419, 2016.

\bibitem{helber2019eurosat}
P.~Helber, B.~Bischke, A.~Dengel, and D.~Borth, ``Eurosat: A novel dataset and deep learning benchmark for land use and land cover classification,'' \emph{IEEE J Sel Top Appl Earth Obs Remote Sens}, vol.~12, no.~7, pp. 2217--2226, 2019.

\bibitem{tschandl2018ham10000}
P.~Tschandl, C.~Rosendahl, and H.~Kittler, ``The ham10000 dataset, a large collection of multi-source dermatoscopic images of common pigmented skin lesions,'' \emph{Sci. Data}, vol.~5, no.~1, pp. 1--9, 2018.

\bibitem{codella2019skin}
N.~Codella, V.~Rotemberg, P.~Tschandl, M.~E. Celebi, S.~Dusza, D.~Gutman, B.~Helba, A.~Kalloo, K.~Liopyris, and M.~Marchetti, ``Skin lesion analysis toward melanoma detection 2018: A challenge hosted by the international skin imaging collaboration (isic),'' \emph{arXiv preprint arXiv:1902.03368}, 2019.

\bibitem{wang2017chestx}
X.~Wang, Y.~Peng, L.~Lu, Z.~Lu, M.~Bagheri, and R.~M. Summers, ``Chestx-ray8: Hospital-scale chest x-ray database and benchmarks on weakly-supervised classification and localization of common thorax diseases,'' in \emph{CVPR}, 2017, pp. 2097--2106.

\bibitem{vinyals2016matching}
O.~Vinyals, C.~Blundell, T.~Lillicrap, and D.~Wierstra, ``Matching networks for one shot learning,'' \emph{NIPS}, vol.~29, 2016.

\bibitem{finn2017model}
C.~Finn, P.~Abbeel, and S.~Levine, ``Model-agnostic meta-learning for fast adaptation of deep networks,'' in \emph{ICML}.\hskip 1em plus 0.5em minus 0.4em\relax PMLR, 2017, pp. 1126--1135.

\bibitem{sung2018learning}
F.~Sung, Y.~Yang, L.~Zhang, T.~Xiang, P.~H. Torr, and T.~M. Hospedales, ``Learning to compare: Relation network for few-shot learning,'' in \emph{CVPR}, 2018, pp. 1199--1208.

\bibitem{snell2017prototypical}
J.~Snell, K.~Swersky, and R.~Zemel, ``Prototypical networks for few-shot learning,'' \emph{NIPS}, vol.~30, 2017.

\bibitem{satorras2018few}
V.~G. Satorras and J.~B. Estrach, ``Few-shot learning with graph neural networks,'' in \emph{ICLR}, 2018.

\bibitem{wang2021cross}
H.~Wang and Z.~Deng, ``Cross-domain few-shot classification via adversarial task augmentation,'' \emph{IJCAI}, 2021.

\bibitem{zheng2023cross}
H.~Zheng, R.~Wang, J.~Liu, and A.~Kanezaki, ``Cross-level distillation and feature denoising for cross-domain few-shot classification,'' in \emph{ICLR}, 2023.

\bibitem{yazdanpanah2022visual}
M.~Yazdanpanah and P.~Moradi, ``Visual domain bridge: A source-free domain adaptation for cross-domain few-shot learning,'' in \emph{CVPR}, 2022, pp. 2868--2877.

\bibitem{oh2022refine}
J.~Oh, S.~Kim, N.~Ho, J.~Kim, H.~Song, and S.~Yun, ``Refine: Re-randomization before fine-tuning for cross-domain few-shot learning,'' in \emph{ACM CIKM}, 2022, pp. 4359--4363.

\bibitem{li2023knowledge}
P.~Li, F.~Liu, L.~Jiao, S.~Li, L.~Li, X.~Liu, and X.~Huang, ``Knowledge transduction for cross-domain few-shot learning,'' \emph{PR}, vol. 141, p. 109652, 2023.

\bibitem{han2022survey}
K.~Han, Y.~Wang, H.~Chen, X.~Chen, J.~Guo, Z.~Liu, Y.~Tang, A.~Xiao, C.~Xu, and Y.~Xu, ``A survey on vision transformer,'' \emph{TPAMI}, vol.~45, no.~1, pp. 87--110, 2022.

\bibitem{hu2022adversarial}
Y.~Hu and A.~J. Ma, ``Adversarial feature augmentation for cross-domain few-shot classification,'' in \emph{ECCV}.\hskip 1em plus 0.5em minus 0.4em\relax Springer, 2022, pp. 20--37.

\bibitem{das2022confess}
D.~Das, S.~Yun, and F.~Porikli, ``Confess: A framework for single source cross-domain few-shot learning,'' in \emph{ICLR}, 2022.

\bibitem{liang2021boosting}
H.~Liang, Q.~Zhang, P.~Dai, and J.~Lu, ``Boosting the generalization capability in cross-domain few-shot learning via noise-enhanced supervised autoencoder,'' in \emph{ICCV}, 2021, pp. 9424--9434.

\bibitem{zhou2023revisiting}
F.~Zhou, P.~Wang, L.~Zhang, W.~Wei, and Y.~Zhang, ``Revisiting prototypical network for cross domain few-shot learning,'' in \emph{CVPR}, 2023, pp. 20\,061--20\,070.

\bibitem{infonce}
A.~v.~d. Oord, Y.~Li, and O.~Vinyals, ``Representation learning with contrastive predictive coding,'' \emph{arXiv preprint arXiv:1807.03748}, 2018.

\bibitem{pattern}
W.~Zhou, S.~Newsam, C.~Li, and Z.~Shao, ``Patternnet: A benchmark dataset for performance evaluation of remote sensing image retrieval,'' \emph{ISPRS journal of photogrammetry and remote sensing}, vol. 145, pp. 197--209, 2018.

\bibitem{jia2022visual}
M.~Jia, L.~Tang, B.~Chen, C.~Cardie, S.~Belongie, B.~Hariharan, and S.~Lim, ``Visual prompt tuning,'' in \emph{ECCV}.\hskip 1em plus 0.5em minus 0.4em\relax Springer, 2022, pp. 709--727.

\end{thebibliography}



\vspace{-25pt}

\begin{IEEEbiography}[{\includegraphics[width=1in,height=1.25in,clip,keepaspectratio]{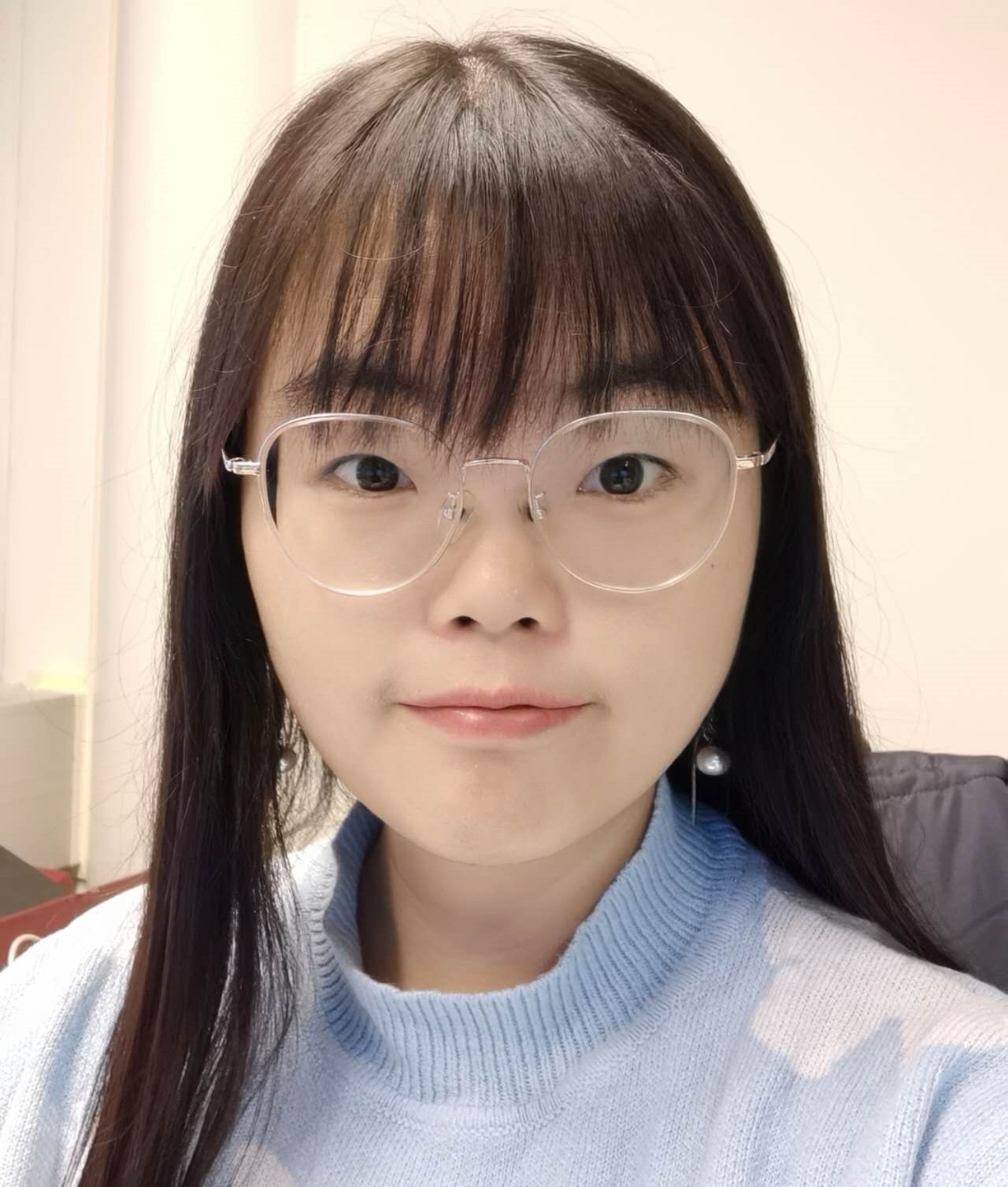}}]{Huali Xu}
 is currently pursuing the Ph.D. degree in Center for Machine Vision and Signal Analysis (CMVS) with the Faculty of Information Technology and Electrical Engineering, University of Oulu, Oulu, Finland. Her current research interests include computer vision, deep learning, few-shot learning, and cross-domain few-shot learning.
\end{IEEEbiography}

\vspace{-25pt}

\begin{IEEEbiography}[{\includegraphics[width=1in,height=1.25in,clip,keepaspectratio]{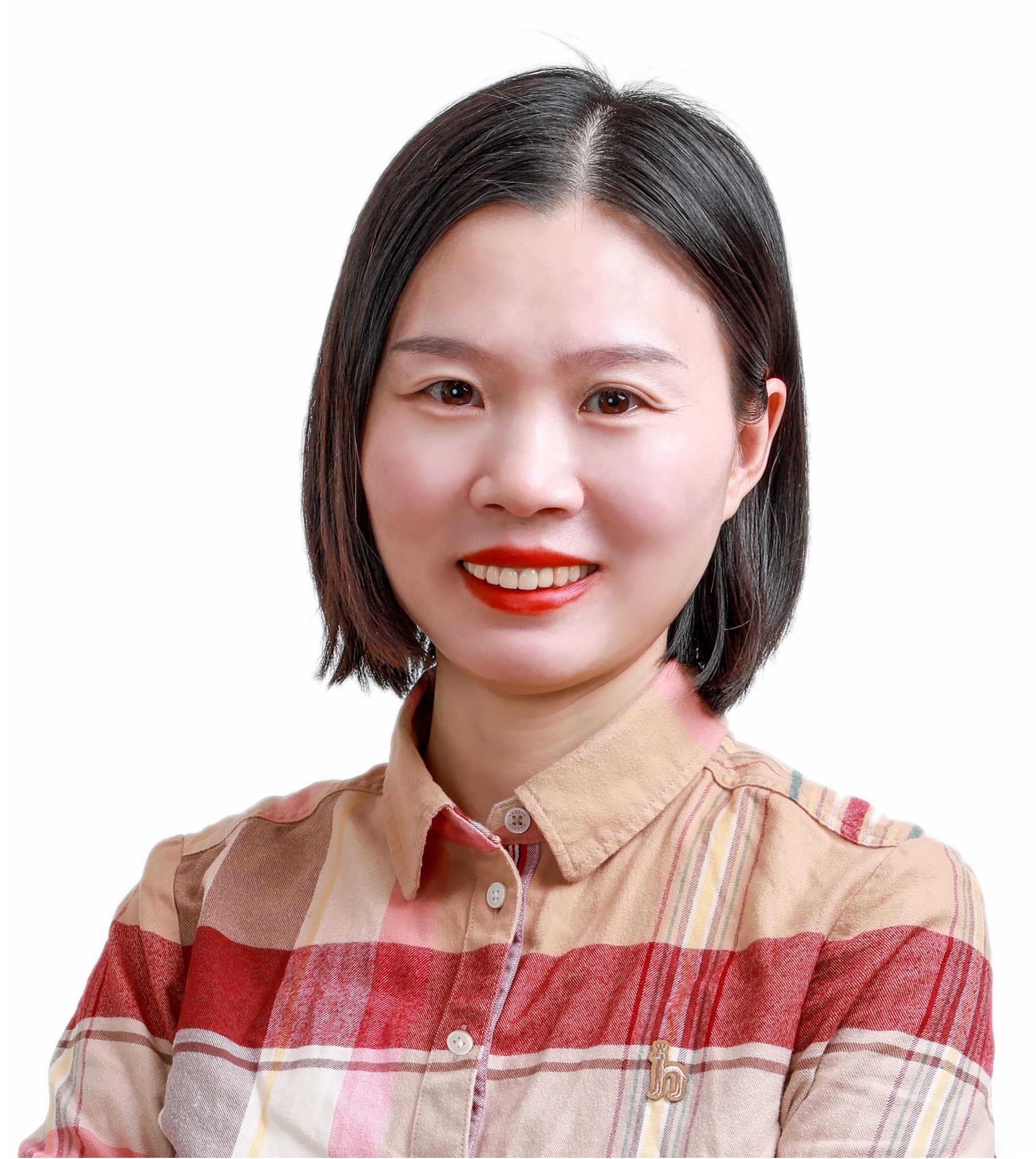}}]{Li Liu}
received the PhD degree in information and communication engineering from the National University of Defense Technology (NUDT), China, in 2012. She is now a full professor with the College of Electronic Science and Technology, NUDT. During her PhD study, she spent more than two years as a visiting student with the University of Waterloo, Canada, from 2008 to 2010. From 2015 to 2016, she spent ten months visiting the Multimedia Laboratory with the Chinese University of Hong Kong. From 2016 to 2018, she worked as a senior researcher with the Machine Vision Group, University of Oulu, Finland. She was a cochair of nine International Workshops with CVPR, ICCV, and ECCV. She served as the leading guest editor for special issues in IEEE TPAMI and IJCV. She is serving as the leading guest editor for IEEE PAMI special issue on “Learning with Fewer Labels in Computer Vision”. Her current research interests include computer vision, pattern recognition and machine learning. Her papers have currently more than 13000 citations according to Google Scholar. She currently serves as associate editor for IEEE TGRS, IEEE TCSVT, and PR.
\end{IEEEbiography}

\vspace{-25pt}

\begin{IEEEbiography}[{\includegraphics[width=1in,height=1.25in,clip,keepaspectratio]{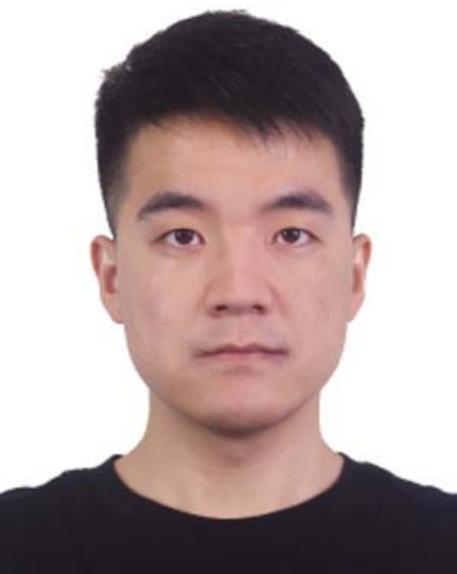}}]{Shuaifeng Zhi}
received his Ph.D. degree in Computing Research at the Dyson Robotics Laboratory, Imperial College London, UK, in 2021. He is currently a Lecturer (Assistant Professor) at the Department of Electronic Science and Technology, National University of Defense Technology (NUDT),
Changsha, China. He was a 6-month visiting student in 5GIC, University of Surrey, UK, in 2015. His current research interests focus on robot vision, particularly on scene understanding, neural scene representation, and semantic SLAM.
\end{IEEEbiography}

\vspace{-25pt}

\begin{IEEEbiography}[{\includegraphics[width=1in,height=1.25in,clip,keepaspectratio]{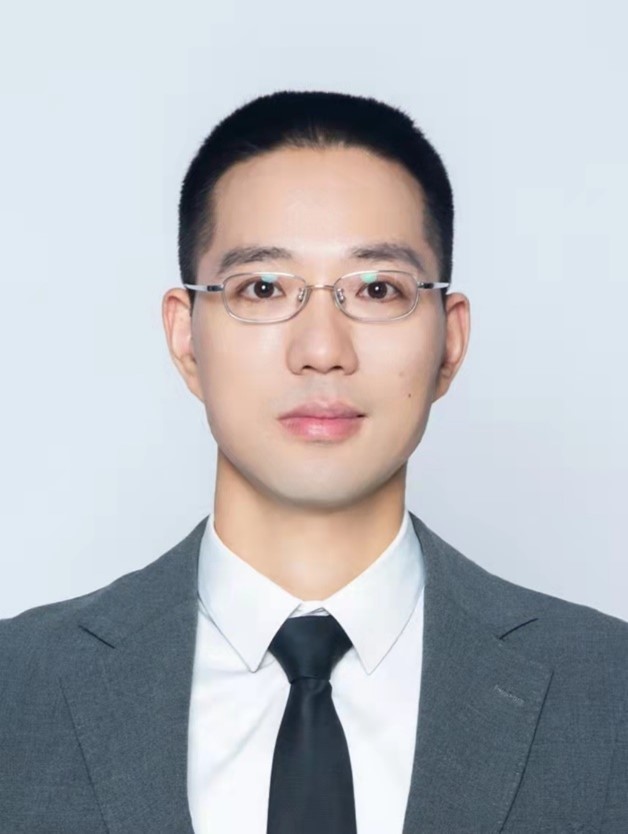}}]{Shaojing Fu}
received the Ph.D. degree in applied mathematicas from the National University of Defense Technology in 2010. He spent a year as a Joint Doctoral Student at the University of Tokyo for one year. He is currently a Professor with the College of Computer, National University of Defense Technology. His research interests include cryptography theory and application, security in cloud, and mobile computing.
\end{IEEEbiography}

\vspace{-25pt}

\begin{IEEEbiography}[{\includegraphics[width=1in,height=1.25in,clip,keepaspectratio]{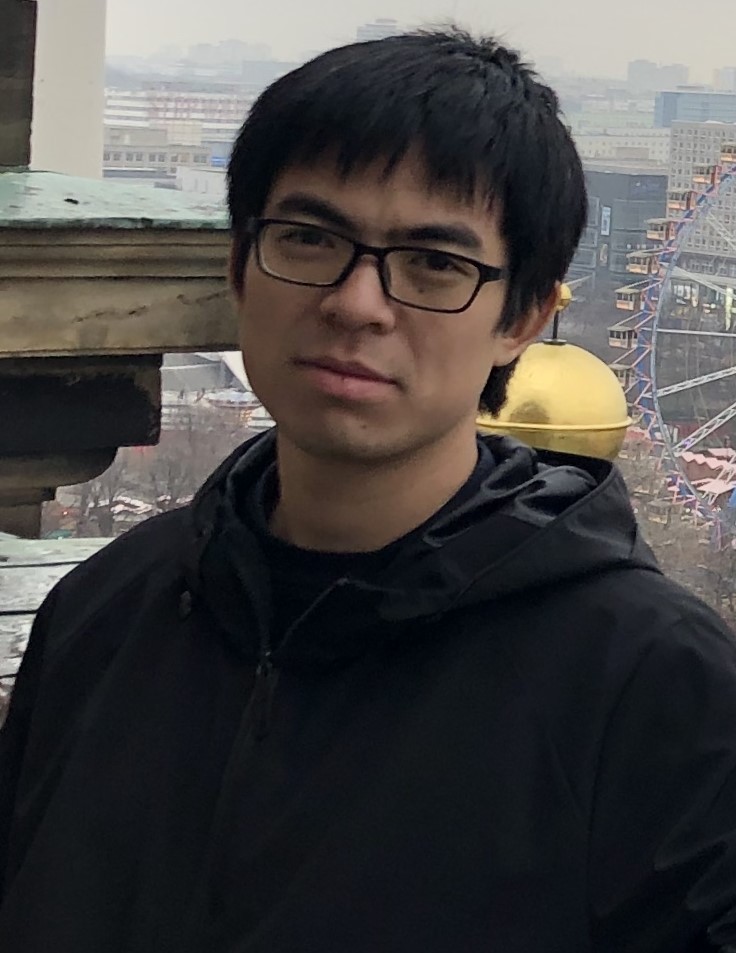}}]{Zhuo Su}
received the doctoral degree in Computer Science and Engineering from University of Oulu, Finland, in 2023, and the B.E. and M.S. degree in School of Automation Science and Electrical Engineering from Beihang University, China, in 2015 and 2018, respectively. He is currently a postdoctoral researcher from the Center for Machine Vision and Signal Analysis, University of Oulu. His research interests include deep neural network compression and acceleration, 3D understanding, and generative AI in computer vision.
\end{IEEEbiography}

\vspace{-25pt}

\begin{IEEEbiography}[{\includegraphics[width=1in,height=1.25in,clip,keepaspectratio]{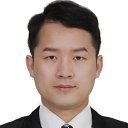}}]{Ming-Ming Cheng}
received his PhD degree from Tsinghua University in 2012. Then he did two years research fellow with Prof. Philip Torr in Oxford. He is now a professor at Nankai University, leading the Media Computing Lab. His research interests include computer graphics, computer vision, and image processing. He received research awards, including ACM China Rising Star Award, IBM Global SUR Award, and CCF-Intel Young Faculty Researcher Program. He is on the editorial boards of IEEE TPAMI/TIP.
\end{IEEEbiography}

\vspace{-25pt}

\begin{IEEEbiography}[{\includegraphics[width=1in,height=1.25in,clip,keepaspectratio]{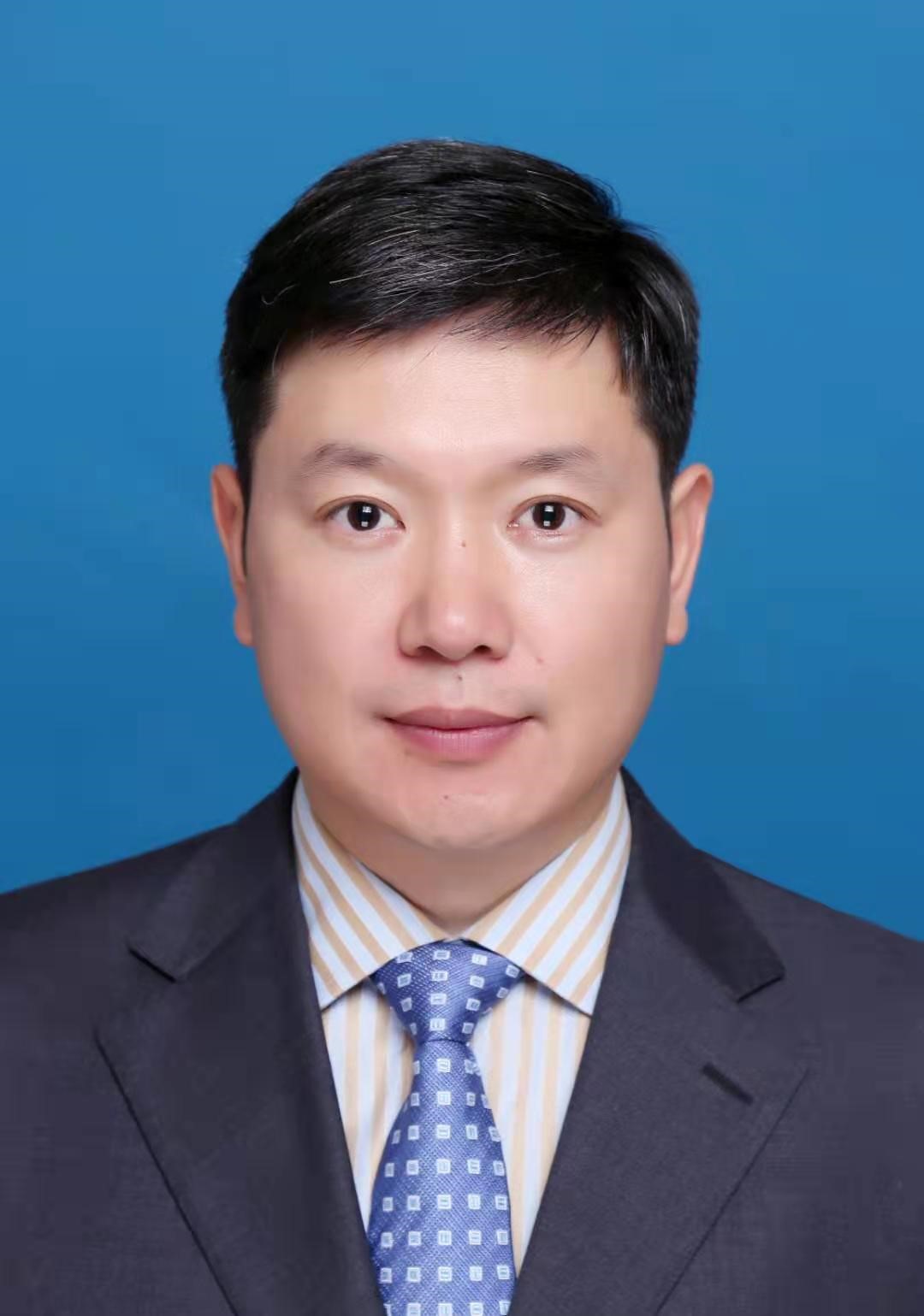}}]{Yongxiang Liu}
received his Ph.D. degree in Information and Communication Engineering from National University of Defense Technology (NUDT), Changsha, China, in 2004. Currently, He is a Full Professor in the College of Electronic Science and Technology, National University of Defense Technology. His research interests mainly include remote sensing imagery analysis, radar signal processing, Synthetic Aperture Radar (SAR) object recognition and Inverse SAR (ISAR) imaging, and machine learning.
\end{IEEEbiography}

\vfill

\end{document}